\documentclass[como,a4paper,12pt,final,nociteindex,noattachments]{classes/c4e-preprint}
 
\usepackage{longtable}
\usepackage{booktabs}
\usepackage{array}
\usepackage{graphicx}
\usepackage[hybrid]{markdown}
\usepackage[utf8]{inputenc}
\usepackage{xcolor}
\usepackage{listings}
\newcolumntype{P}[1]{>{\raggedright\arraybackslash}p{#1}}

\usepackage[utf8]{inputenc} 
\usepackage[T1]{fontenc}    

\makeatletter

\@ifpackageloaded{xcolor}
{\relax}
{%
\usepackage[dvipsnames]{xcolor}
}%

\usepackage{xspace}

\usepackage{amsfonts, amsmath, amssymb, amscd, amsthm, bm, array}

\@ifpackageloaded{mhchem}
{\relax}
{%
\usepackage[version=4]{mhchem} 
}%

\@ifpackageloaded{xr-hyper}
{\relax}
{%
\usepackage{xr-hyper}
}%

\@ifpackageloaded{hyperref}
{\relax}
{%
\usepackage[hyphens]{xurl}
\usepackage[hidelinks]{hyperref}
}%
\usepackage[intoc]{nomencl}
\makenomenclature

\@ifpackageloaded{enumitem}
{\relax}
{%
\usepackage[shortlabels]{enumitem}
}%

\@ifpackageloaded{ulem}
{\relax}
{%
\usepackage[normalem]{ulem}
}%
\makeatother

\usepackage{lineno}
\usepackage{xspace}

\usepackage{listings}
\makeatletter
\newcommand*{\addFileDependency}[1]{
    \typeout{(#1)}
    \@addtofilelist{#1}
    \IfFileExists{#1}{}{\typeout{No file #1.}}
}
\makeatother

\usepackage{subfigure}
\usepackage{wrapfig}
\usepackage{placeins}
\usepackage{booktabs}
\usepackage{multirow}
\usepackage{tabularx,xtab}
\makeatletter
\@ifpackageloaded{algorithm}
{\relax}
{%
\usepackage[linesnumbered,ruled,vlined]{algorithm2e}
}%
\makeatother

\newlength{\myfigheight}
\newlength{\myfigwidth}
\newlength{\mygutwidth}
\makeatletter
\newcommand\appendtographicspath[1]{%
  \ifx\Ginput@path\@undefined
    \graphicspath{#1}%
  \else
    \g@addto@macro\Ginput@path{#1}%
  \fi
}
\makeatother
\appendtographicspath{{./}{figures/}}

\newcommand{\ie}[0]{\textit{i.e\samesentenceperiod}\xspace} 
\newcommand{\eg}[0]{\textit{e.g\samesentenceperiod}\xspace} 
\newcommand{\vs}[0]{\textit{vs}\xspace} 
\makeatletter
\AtBeginDocument{%
  \renewcommand{\eg}{\textit{e.g.}\xspace}%
  \renewcommand{\ie}{\textit{i.e.}\xspace}%
}
\makeatother

\makeatletter
\@ifpackageloaded{listings}
{\relax}
{%
\@ifpackageloaded{minted}{}{
    \usepackage{caption}
    \usepackage[newfloat]{minted}
    \captionsetup[listing]{position=top}
}
}%
\makeatother

\newcommand{\addressceb}{
    Department of Chemical Engineering\\and Biotechnology\\
    University of Cambridge\\
    Philippa Fawcett Drive\\
    Cambridge, CB3 0AS\\
    United Kingdom
    \vspace{2ex} 
}
\newcommand{\addresscares}{
    CARES\\
    Cambridge Centre for Advanced\\Research and Education in Singapore\\
    1 Create Way\\
    CREATE Tower, \#05-05\\
    Singapore, 138602
}

\newcommand{\addresscmcl}{
    CMCL\\
    No. 9, Journey Campus\\
    Castle Park\\
    Cambridge\\
    CB3 0AX\\
    United Kingdom
}
\newcommand{\addresscmpg}{
    CMPG\\
    GRIPS -- Gründerinnenzentrum Pirmasens\\
    Delaware Avenue 1--3\\
    66953 Pirmasens\\
    Germany
}

\newcommand{\addressmit}{
    MIT, Chemical Engineering\\
    77 Massachusetts Avenue, Room E17-504\\
    Cambridge, MA  02139  USA
}

\newcommand{\sectionprefix}{}

\providecommand{\doi}[1]{\href{https://doi.org/#1}{doi:#1}}

\makeatletter
\renewcommand*{\addFileDependency}[1]{
	\typeout{(#1)}
	\@addtofilelist{#1}
	\IfFileExists{#1}{}{\typeout{No file #1.}}
}
\makeatother

\usepackage{silence}
\usepackage{xspace}

\makeatletter
\AtBeginDocument{%
  \renewcommand{\eg}{\textit{e.g.}\xspace}%
  \renewcommand{\ie}{\textit{i.e.}\xspace}%
}
\makeatother
\begin{document}


\title{Ontology-to-tools compilation for executable semantic constraint enforcement in LLM agents\\
}


\addauthor{1}{Xiaochi~Zhou} 
\addauthor{1}{Patrick~Butler}
\addauthor{3}{Changxuan~Yang}
\addauthor{5}{Simon~Rihm}
\addauthor{1}{Thitikarn~Angkanaporn}
\addauthor{1,2,4}{Jethro~Akroyd} 
\addauthor{1,2,4}{Sebastian~Mosbach} 
\addauthor{1,2,3,5}{Markus~Kraft} 

\affiliation{1}{\addressceb}
\affiliation{2}{\addresscares}
\affiliation{3}{\addressmit}
\affiliation{4}{\addresscmcl}
\affiliation{5}{\addresscmpg}

\addsubject{} 

\keywords{Large language models; autonomous agents; knowledge graphs; scientific information extraction}
\nopreprint{343}  \releasedate{\today}

\setpdfmetadata
\maketitle


\begin{abstract}
    \noindent%

We introduce \emph{ontology-to-tools compilation} as a proof-of-principle mechanism for coupling large language models (LLMs) with formal domain knowledge. Within \emph{The World Avatar} (TWA), ontological specifications are compiled into executable tool interfaces that LLM-based agents must use to create and modify knowledge graph instances, enforcing semantic constraints during generation rather than through post-hoc validation. Extending TWA’s semantic agent composition framework, the Model Context Protocol (MCP) and associated agents are integral components of the knowledge graph ecosystem, enabling structured interaction between generative models, symbolic constraints, and external resources. An agent-based workflow translates ontologies into ontology-aware tools and iteratively applies them to extract, validate, and repair structured knowledge from unstructured scientific text. Using metal--organic polyhedra synthesis literature as an illustrative case, we show how executable ontological semantics can guide LLM behaviour and reduce manual schema and prompt engineering, establishing a general paradigm for embedding formal knowledge into generative systems.

    \highlights{include-highlights.tex}
\end{abstract}

\clearpage \setcounter{tocdepth}{3} \tableofcontents \clearpage

\section{Introduction}\label{sec:introduction}


Extracting structured data from scientific literature is a critical and routine step for enabling machine-actionable scientific workflows, including hypothesis generation, data-driven modelling, and synthesis prediction.~\cite{grishman2015,krallinger2017,xu_large_2024}. In reticular chemistry, for example, literature extraction can support synthesis predictions of novel materials and estimation of properties such as gas adsorption, and provide means to identify gaps in the immediate chemical space~\cite{theworldavatar2025,kondinski2022,zheng2023,jablonka2024}. Moreover, structuring literature makes it possible to link extracted records to complementary data sources, expanding their context and increasing their value beyond what is reported in any single paper~\cite{martinezrodriguez2020,liang_survey_2024}.


However, turning scientific text into machine-actionable data remains challenging. Scientific text requires precise interpretation and is domain-specific, while downstream uses, such as data-driven modelling, require consistently defined records that are comprehensively validated~\cite{grishman2015,krallinger2017,xu_large_2024,schilling-wilhelmi_text_2025}. These records need clear roles and concept boundaries, normalized units and values, and, crucially, unambiguous entity references~\cite{wimalasuriya2010,martinezrodriguez2020,schilling-wilhelmi_text_2025}. For example, a reagent may be mentioned without an explicit role (precursor \vs \ solvent), experimental conditions may be expressed in inconsistent unit forms (\eg C \vs Degrees Celsius), and the same chemical may appear under multiple names or abbreviations; downstream workflows therefore require canonical identifiers (\eg, CAS numbers or InChIKeys) and consistent typing to support validation, deduplication, and cross-paper integration~\cite{pascazio2023,rihm2025}.


Ontologies are one natural option for specifying these roles, boundaries, and constraints: an ontology is a formal, machine-readable description of a domain, typically decomposed into a T-Box that defines concepts and relations, and an A-Box that instantiates them with concrete entities and facts~\cite{gruber1993,guarino2009}. In such knowledge graphs, entities are represented by Internationalized Resource Identifiers (IRIs), which serve as unique identifiers for individual instances, such as specific chemical species. In parallel, recent large language models (LLMs) have enabled few-shot and zero-shot adaptation for extraction tasks, and can apply prompt-based, soft definitions of roles and boundaries by interpreting context and producing schema-shaped records~\cite{xu_large_2024}.


Pipelines that aim to produce structured data ready for downstream use often combine ontologies (to define target roles, relations, and constraints) with extractors (LLMs or otherwise) to interpret text. However, downstream use typically requires that extracted records conform to the ontology, including valid types and relations, normalized values and units, and grounded identifiers. This concentrates effort in constraint enforcement, such as validation, normalization, and grounding or alignment, implemented within the extraction system or as a separate step~\cite{wimalasuriya2010,martinezrodriguez2020,xu_large_2024}. In non-LLM pipelines, this enforcement is commonly realized through domain-specific components such as trained models, hand-engineered rules, and ontology-aligned assembly procedures~\cite{wimalasuriya2010,martinezrodriguez2020}.


Many other approaches shift this work to post-hoc processing: they first generate schema-shaped records, then apply deterministic validation and ontology/database alignment to ensure the outputs satisfy required fields, normalization, cross-field consistency, and identifier grounding~\cite{xu_large_2024,schilling-wilhelmi_text_2025,liu_we_2024,xu2023}. As a result, achieving ontology-conformant data often depends on hand-built validation/alignment logic outside the extractor, rather than being enforced during extraction itself.

Across these families, the common limitation is how downstream requirements are enforced: they are implemented either as bespoke, domain-specific pipeline logic around extraction or as a separate post-hoc layer for validation, normalization, and grounding and alignment, concentrating expert effort in code that must be revised as schemas and scope change.


Recently, large language models (LLMs) have increasingly supported tool calling, where a model invokes external functions (scripts and APIs) during generation to produce structured outputs, perform deterministic checks, and retrieve supporting evidence. Frameworks such as the Model Context Protocol (MCP) standardize this interaction by providing a common interface for registering tools and exchanging typed inputs and outputs~\cite{openai2024,anthropic2024mcp,mcp2025spec,yao2023react,schick_toolformer_2023,wu_autogen_2023,liang_survey_2024,ren2024survey,liu_we_2024,li_large_2024}. These capabilities make it possible to enforce structure and domain constraints during extraction, rather than relying solely on post hoc validation.

Such constraints are particularly important in scientific domains, where extracted information must align with well-defined ontologies and existing knowledge bases. The World Avatar~\cite{theworldavatar2025} is a large-scale dynamic knowledge graph for chemistry that provides curated T-Boxes and extensive A-Box instances covering synthesis descriptions, metal organic polyhedra, and chemical species~\cite{pascazio2023,kondinski2022,rihm2025}. As a result, it offers a natural grounding target for tool-augmented LLM extraction, supplying ready-to-use semantic schemas against which generated outputs can be validated and instantiated.



Motivated by the constraint-enforcement bottleneck identified above, we introduce a novel ontology-to-tools compilation framework that turns an ontology into executable LLM-callable tools, enabling constraints to be enforced during extraction rather than by bespoke pipeline logic or post-hoc validation. To our knowledge, no prior system has transformed an ontology into an LLM-executable constraint enforcement layer. In our approach, an LLM-driven agent performs document-level extraction and  invoking compiled tools to construct the knowledge graph directly: tool calls instantiate individuals and relations with built-in constraint checks and structured feedback on violations. The same compilation framework also yields lexical grounding tools. These tools link surface text mentions to ontology-defined entities using lexical labels and evidence from a reference knowledge graph. In practice, they align extracted mentions to the IRIs of existing instances in the graph, for example an already recorded chemical species.

Our contribution is: (1) an ontology-to-tools compilation framework for generating
ontology-aligned prompts and MCP tools from a \emph{T-Box} and meta-prompts; (2)
an ontology-constrained KG construction procedure that enforces constraints at
creation time via tool calls with structured feedback; and (3) an ontology-driven grounding workflow that generates lexical grounding tools from the \emph{T-Box} and endpoint evidence for identifier alignment. We demonstrate the ontology-to-tools compilation framework on metal--organic polyhedra (MOP) synthesis literature in The World Avatar, covering ontology-constrained knowledge-graph construction from synthesis text, followed by lexical grounding to canonical identifiers.

From a machine-intelligence perspective, the central contribution is a compilation mechanism that transforms symbolic constraints into executable action interfaces for generative models. This reframes constraint enforcement from prompt- or schema-based output constraints and post-hoc validation into run-time interaction with a structured environment. The resulting behaviour is not achieved through prompt engineering or grammar restriction, but through tool-mediated interaction with a persistent symbolic state. The chemistry case study serves as a concrete instantiation of this general mechanism.



\section{Knowledge-Graph Construction from Metal--Organic Polyhedra Synthesis Literature}\label{sec:application}
\begin{figure}
    \centering
    \includegraphics[width=1\linewidth]{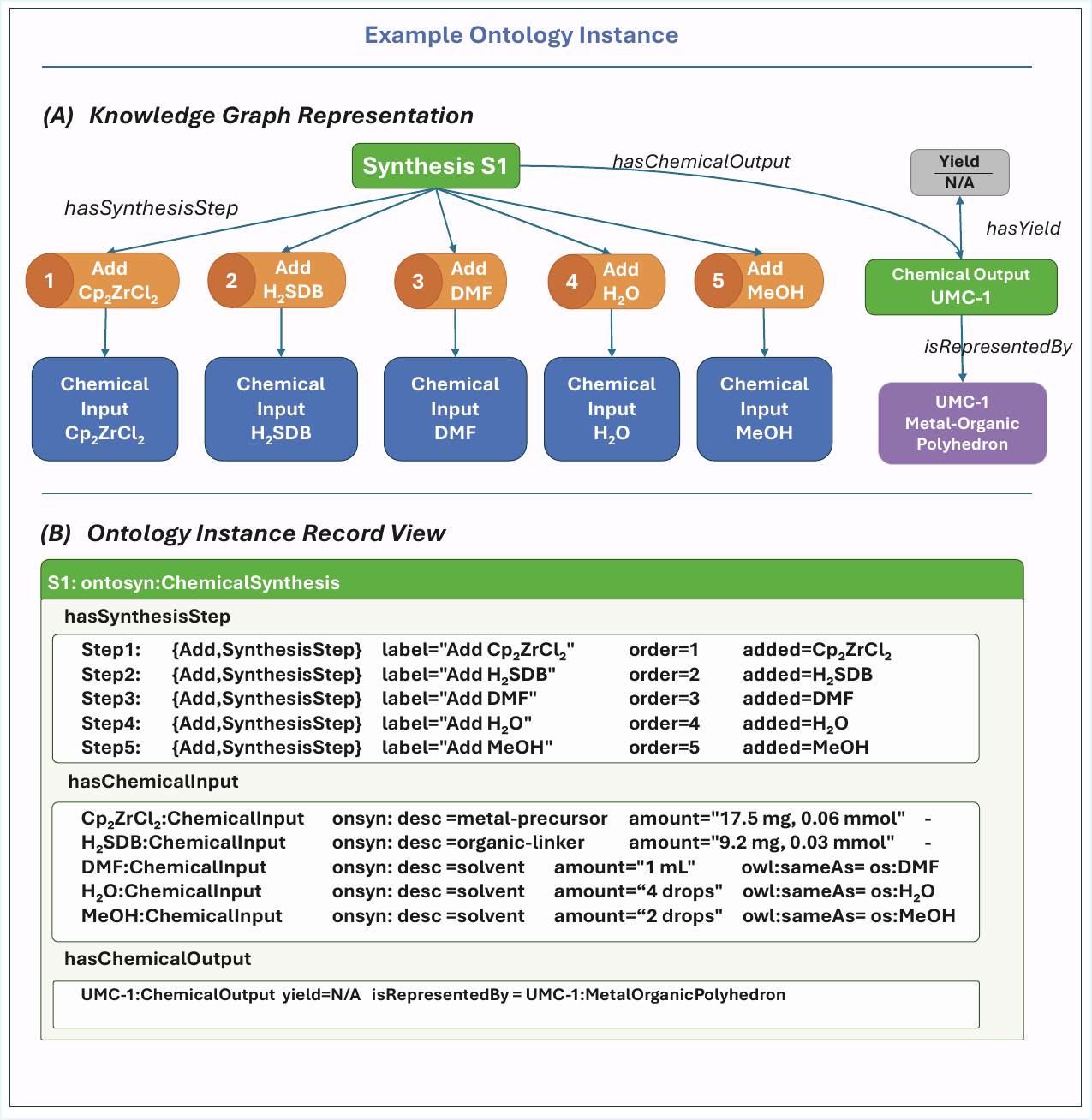}
\caption{\textbf{Example ontology instance produced by the instantiation agent.}
(A) A-Box subgraph instantiated under OntoSyn/OntoMOPs for synthesis \texttt{S1}, including ordered synthesis steps, chemical inputs, and product \texttt{UMC-1} (with yield and representation links).
(B) Compact record projection (A-Box normal form) of the same instance in canonical slot--value form; \texttt{onsyn:*} abbreviates \texttt{ontosyn:*}, and selected \texttt{ChemicalInput} entities are grounded via \texttt{owl:sameAs} to \texttt{ontospecies:Species}.}
    \label{fig:example_instance}
\end{figure}

We demonstrate the ontology-to-tool compilation framework by constructing grounded, ontology-consistent knowledge graphs from metal–organic polyhedra synthesis literature.
Figure~\ref{fig:example_instance} illustrates the core knowledge-graph representation produced for metal--organic polyhedra (MOP) synthesis papers. The results reported in this section are obtained from 30 MOP-related publications, with each paper treated as a full-text input including manuscript text, tables, and supplementary information when available.

The task definition is to convert unstructured MOP synthesis literature into grounded, ontology-consistent knowledge-graph instances. Given a full-text synthesis paper, the output is a set of interlinked instances capturing synthesis procedures, chemical reagents, and the reported MOP, with extracted entities linked to existing knowledge graph instances. 

The representation draws on multiple complementary domain ontologies, including OntoSynthesis~\cite{rihm2025} for modelling synthesis events, ordered steps, conditions, and reagent usage; OntoSpecies~\cite{pascazio2023} for canonical chemical species identities and identifiers; OntoMOPs~\cite{kondinski2022} for MOP products, structural concepts, and derived chemical building units (CBUs); and OM-2 for representing and normalizing quantities and units in synthesis descriptions. Chemical species are grounded at the usage-instance level by linking extracted instances to canonical OntoSpecies entries.

This task is challenging because it requires the coordinated use and alignment of multiple ontologies when analysing a single paper. Procedural knowledge, canonical chemical identity, and product- and structure-level material knowledge, including derived CBUs, must be populated consistently and linked across ontology boundaries. Moreover, the relevant information is often reported at different levels of detail, and the resulting knowledge graph must keep track of where each piece of information comes from, so that roles, amounts, and other qualifiers remain correctly associated with the appropriate chemical usage and synthesis steps. In addition, CBU derivation requires connecting extracted synthesis evidence to product representations and external chemical or crystallographic data while maintaining consistent identifiers across all ontological layers.

For each paper, the framework produces three interconnected outputs.  
First, a structured synthesis recipe captures ordered synthesis steps, step-level actions, and associated conditions, together with additional synthesis-related details such as reagent usage and provenance information.

Second, a set of grounded chemical species instances represents chemical entities uniquely, with contextual qualifiers such as role, amount, and units attached to each occurrence in the synthesis, and with links to canonical OntoSpecies records for cross-paper integration.

Third, derived chemical building units (CBUs), including reusable metal nodes or clusters and organic ligands, are obtained by combining extracted evidence with database-backed chemical and crystallographic data. These CBUs are instantiated under OntoMOPs and linked back to the reported MOP product and the synthesis in which they were produced.

Figure~\ref{fig:example_instance} shows an example ontology instance produced for a representative synthesis. Panel~(A) presents an excerpt of the instances subgraph, \ie A-Box, including a synthesis event linked to an ordered sequence of steps, step-scoped reagent usage, and the reported MOP product, together with attributes such as yield and representation links. Panel~(B) shows a compact, record-style projection of the same instances in canonical slot--value form used for inspection and downstream querying. In this representation, contextual qualifiers remain attached to individual usage instances, while selected inputs are grounded to canonical OntoSpecies entries via identity links. Together, the outputs integrate procedural structure, species grounding, and product- and CBU-level semantics to yield a single, queryable knowledge-graph instance.


\section{Design of Ontology-to-Tools Compilation Framework}

\begin{figure}
    \centering
    \includegraphics[width=1\linewidth]{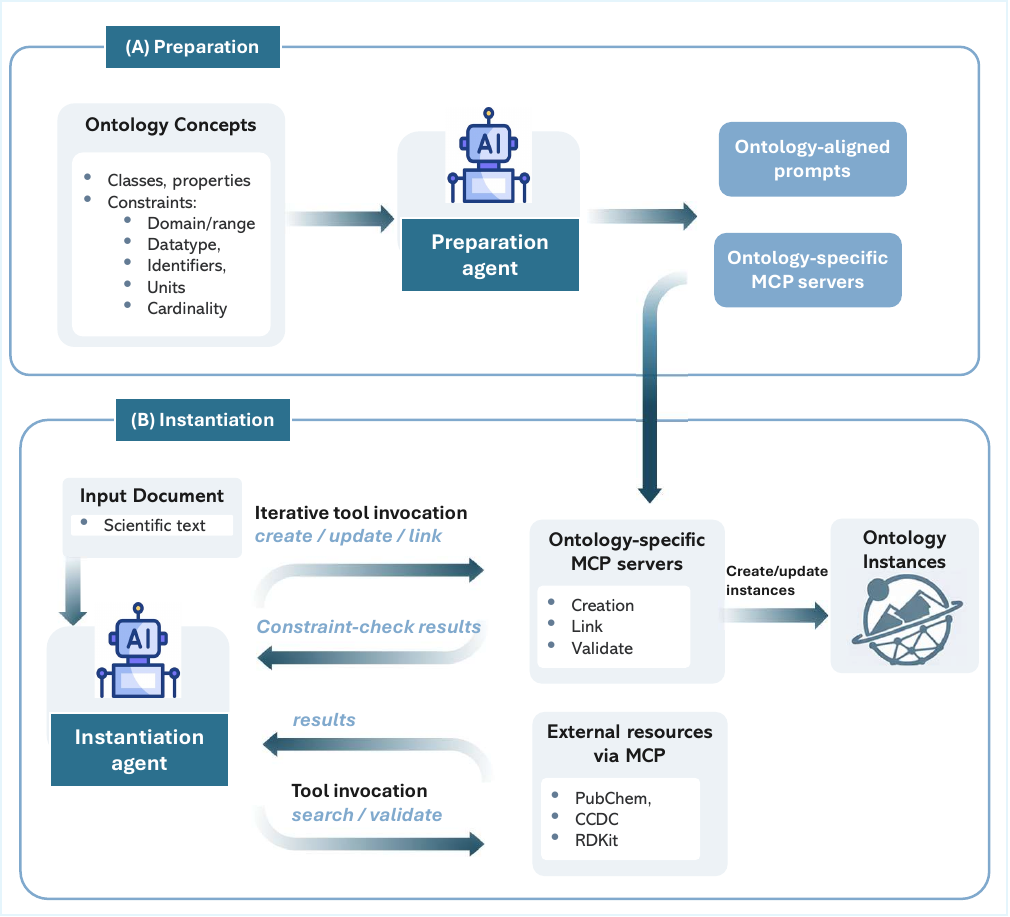}
    \caption{ Ontology-to-tools compilation as an executable semantic control layer for LLM-based agents. Symbolic ontological definitions (T-Box) within The World Avatar are compiled into executable tool interfaces and validators that define the action space available to a large language model during generation. Rather than producing free-form text, the LLM interacts with a persistent symbolic state by invoking ontology-aligned actions that create, modify, and validate graph instances. Constraint violations trigger structured feedback, enabling iterative repair and grounding to external resources. This reframes semantic constraint enforcement from post-hoc validation or constrained decoding into run-time interaction with an evolving symbolic environment, allowing the model to operate as a stateful, ontology-aware agent.}
    \label{fig:workflow}
\end{figure}

This section describes how the ontology-consistent knowledge-graph instances defined in Section~\ref{sec:application} are constructed from unstructured documents. The focus here is on the compilation and execution mechanisms that translate ontology specifications into executable tools and use them to perform constrained extraction. Figure~\ref{fig:workflow} provides an overview of the end-to-end workflow.
The compiled tool interfaces collectively define the action space available to the LLM agent, constraining generation through executable semantics rather prompt- or schema-based constraints on the output.

As illustrated in Figure~\ref{fig:workflow}, the framework has two stages: (1) the \emph{preparation stage}, which compiles an ontology into tools and prompts, and (2) the \emph{instantiation stage}, which runs a tool-using agent to construct the knowledge graph from papers. In the \emph{preparation stage}, the preparation agent consumes an ontology schema (T-Box) together with domain-agnostic meta-prompts (instruction templates) and generates ontology-aligned runtime prompts, supporting scripts, and LLM-callable tool interfaces exposed via the Model Context Protocol (MCP). These tools implement ontology-aware instance construction with built-in validation logic. In the \emph{instantiation stage}, the instantiation agent applies the generated prompts, scripts, and LLM-callable tools to each document, invoking tools to create and link individuals and relations; constraint violations are returned as structured feedback to support iterative completion and repair.

The compilation layer treats the T-Box as a machine-readable contract. It specifies which classes, relations, attributes, and constraints are allowed. From this contract, the framework generates executable tool interfaces with explicitly specified inputs, outputs, and validation behaviour. These tools are the only way to create or modify structured instances, so constraints are checked and repaired during construction rather than only after the fact.

During instantiation, the extraction agent interprets document content and issues tool calls to construct synthesis steps, usage instances, species links, and product entities. External chemical and crystallographic resources are integrated as callable tools within the same execution framework, enabling canonical species identification and database-backed derivation of higher-level entities such as CBUs. All extracted, grounded, and derived entities are instantiated into a knowledge graph under the constraints defined by the ontology contract. Together, these components implement an end-to-end workflow in which ontology specifications and unstructured literature jointly drive the construction of grounded, internally consistent knowledge-graph instances, without relying on post-hoc validation or manual schema-specific repair.


\section{Performance of Ontology-to-tool compilation}\label{sec:results}
  
This section evaluates the ontology-to-tool compilation framework and the resulting knowledge-graph construction on the same curated corpus of 30 metal--organic polyhedra (MOP) synthesis articles used throughout this work. Two questions matter. First, are the generated graphs semantically healthy, \ie do they satisfy the ontology constraints and remain structurally consistent under instantiation? Second, is the extracted and grounded content accurate with respect to the source literature?

\begin{figure}
    \centering
    \includegraphics[width=0.9\linewidth]{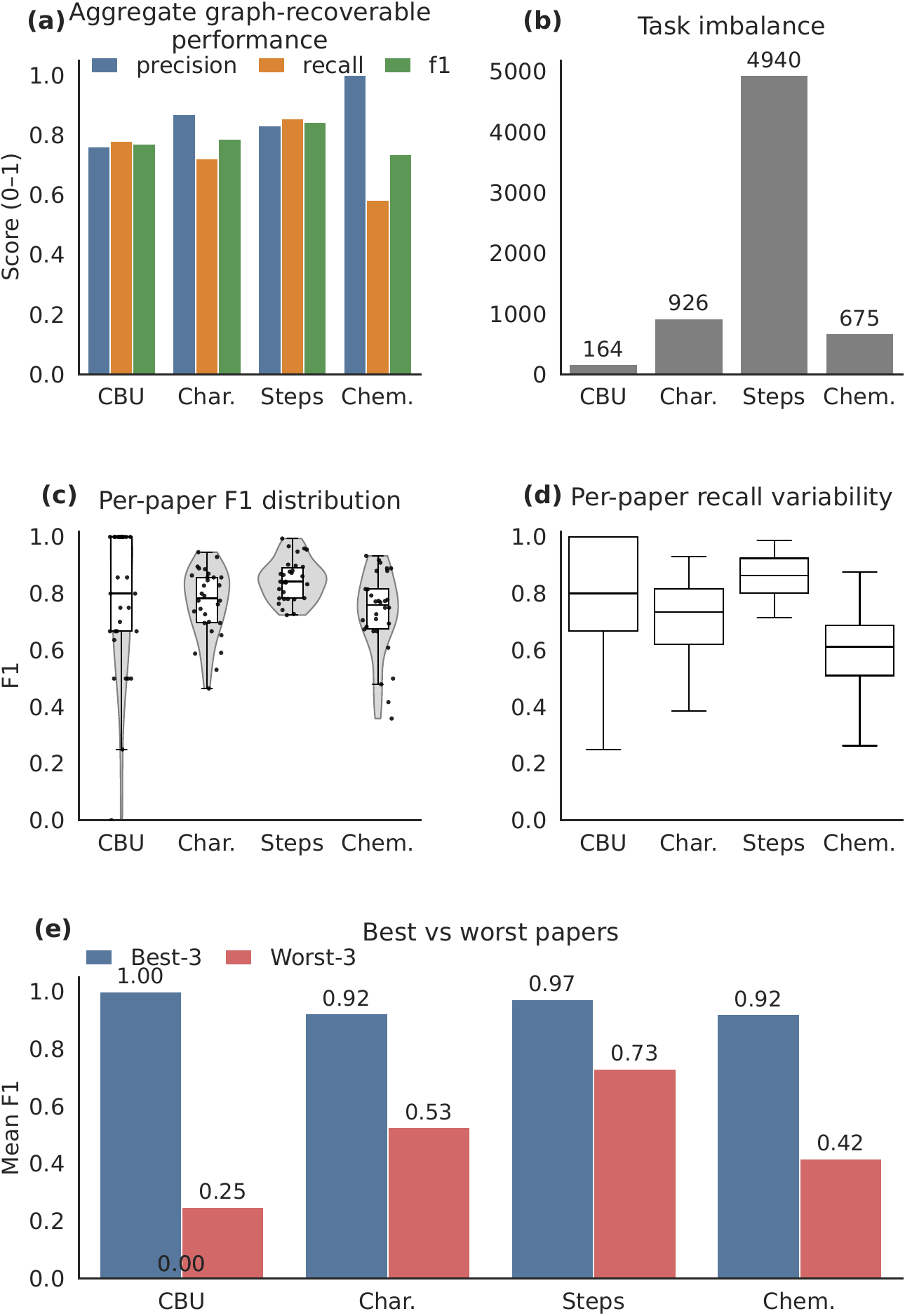}
    \caption{\textbf{Core end-to-end results across four extraction/instantiation domains.}
    (a) Aggregate graph-recoverable precision, recall and F1 for grounded/derived CBUs, characterisation entities, synthesis steps and reaction chemicals (Table~\ref{tab:overall}). 
    (b) Task imbalance in ground-truth positives (Table~\ref{tab:benchmark_profile}). 
    (c) Per-paper F1 distributions across the 30-paper benchmark. 
    (d) Per-paper recall variability, highlighting recall-limited categories. 
    (e) Best--worst paper contrast (mean F1 over top-3 vs bottom-3 papers) summarising dataset heterogeneity.}
    \label{fig:fig1_core_results}
\end{figure}

\begin{figure}
    \centering
    \includegraphics[width=\linewidth]{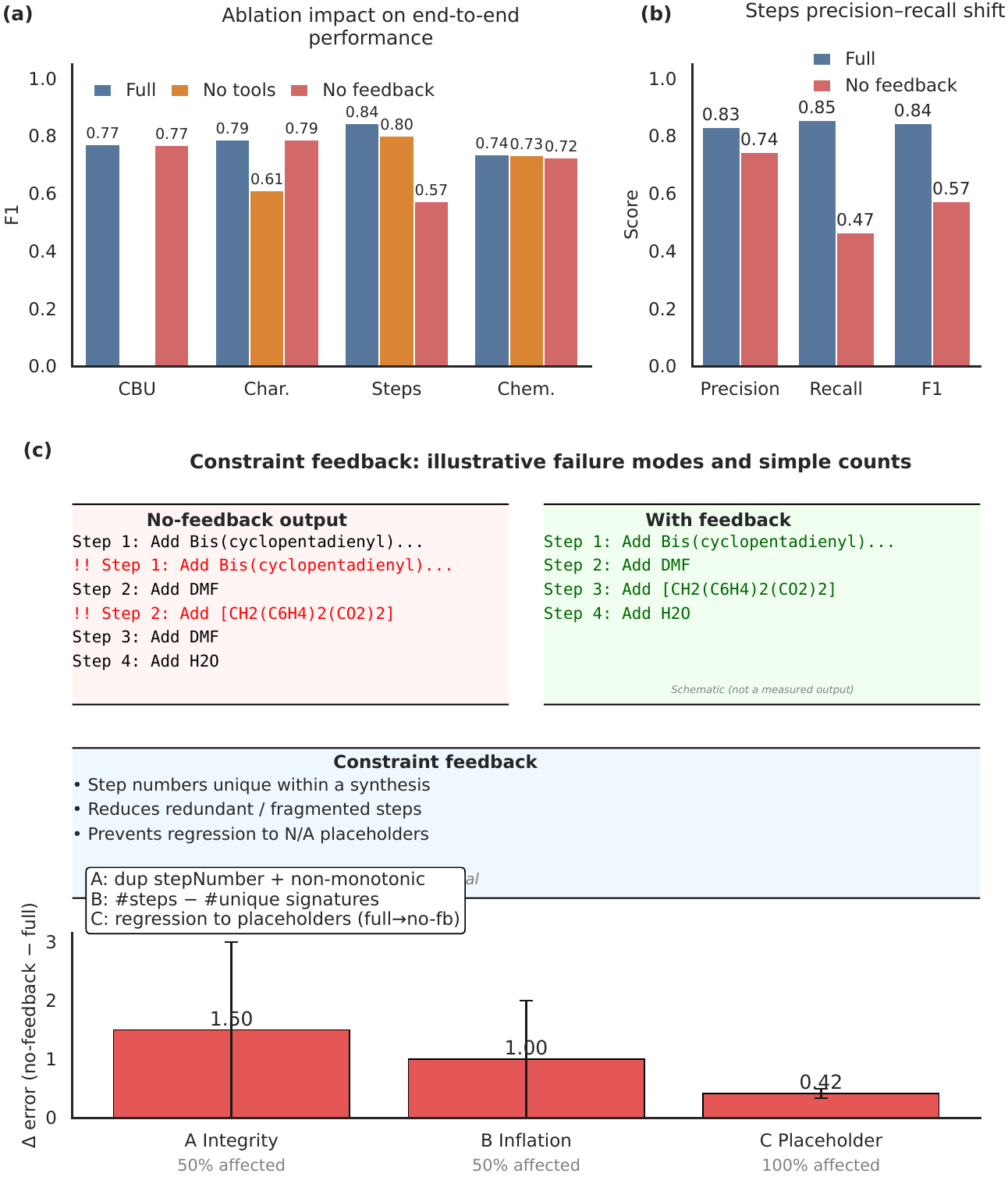}
    \caption{\textbf{Component necessity and the role of constraint feedback.}
    (a) Ablation impact on end-to-end F1 by category (Table~\ref{tab:ablation}). 
    (b) Steps-only precision--recall--F1 shift under feedback removal, illustrating the dominant effect on step completeness/recoverability.
    (c) Constraint feedback: illustrative failure modes and paired $\Delta$ error counts computed over aligned full vs no-feedback syntheses. The excerpt shows typical no-feedback degradations (\eg step-number integrity and redundancy), while the mini-plot quantifies three paired error proxies (A--C; defined in-panel) with bootstrap confidence intervals.}
    \label{fig:fig2_ablation_constraint}
\end{figure}

\begin{figure}
    \centering
    \includegraphics[width=1\linewidth]{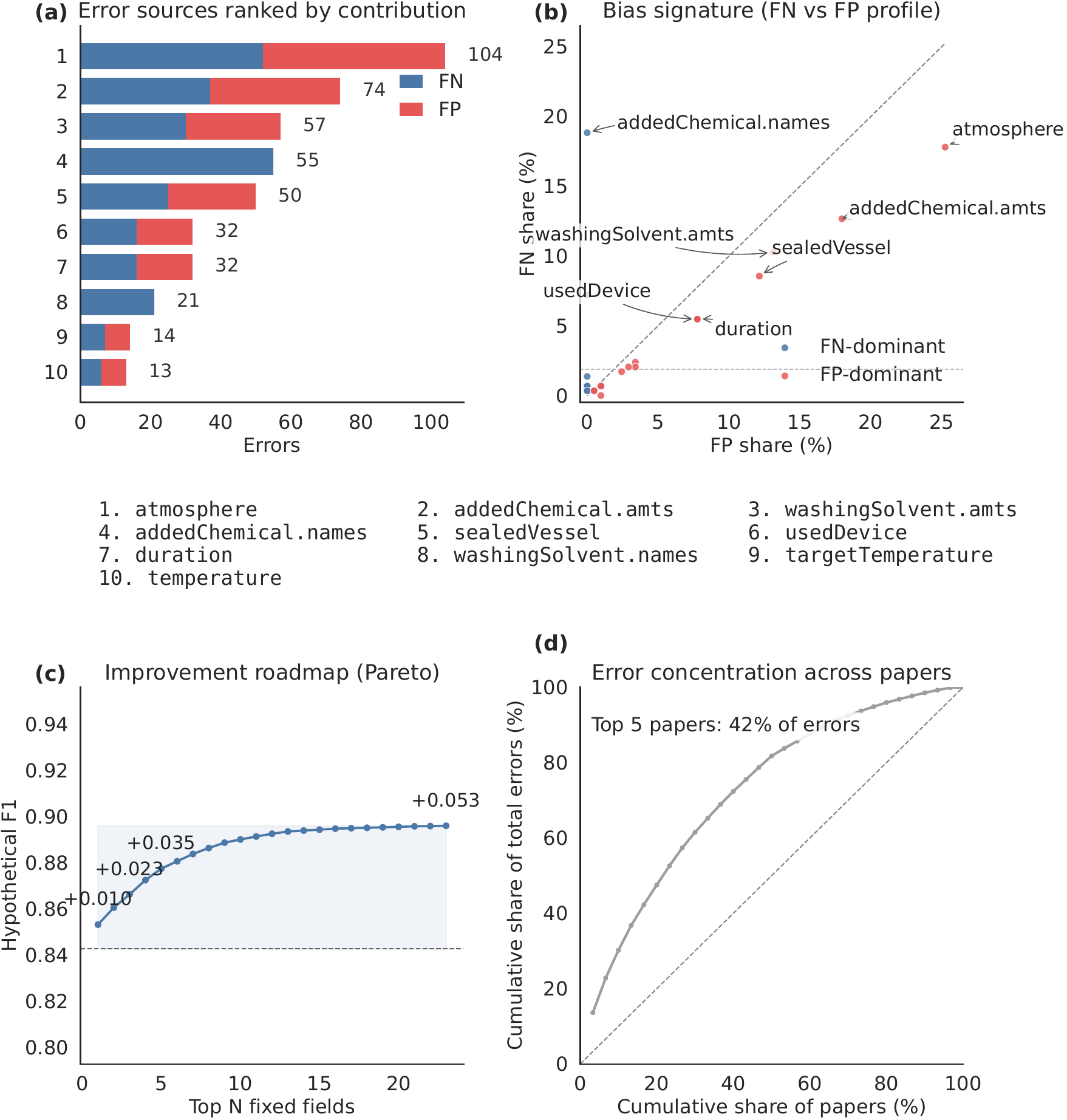}
    \caption{\textbf{Error anatomy and improvement priorities for synthesis steps.}
    (a) Top error-contributing fields (FP vs FN) aggregated over the benchmark. 
    (b) Field-level bias signature (FN-share vs FP-share), separating recall-limited from precision-limited fields.
    (c) Hypothetical improvement roadmap (Pareto): cumulative F1 if the top-$N$ error-contributing fields were corrected.
    (d) Error concentration across papers (Lorenz-style curve), showing whether a small subset of papers accounts for a disproportionate share of step errors.}
    \label{fig:fig3_error_anatomy}
\end{figure}

To answer these questions, we compare the generated outputs against manual ground-truth annotations in predefined JSON record formats covering four target categories: grounded and derived CBUs, characterisation entities, synthesis steps, and reaction chemicals. Section~\ref{subsec:dataset} describes the dataset in detail. We report \emph{graph-recoverable} performance first. This measures whether the information required by the evaluation schema is actually present in the constructed knowledge graph in the expected ontology form, \ie as individuals and relations that can be retrieved deterministically. Concretely, for each target JSON record type, we use a fixed SPARQL query to reconstruct the record from the graph, and we score the reconstructed records against the ground truth. SPARQL querying provides the recoverability test because it requires the relevant facts to be encoded as explicit triples with the correct links between entities, rather than appearing only as free text, partial attributes, or disconnected nodes. The SPARQL queries are fixed in advance and are not adapted per paper or per predicted graph. They are derived from the target ontological schema (T-Box) and the corresponding JSON record definitions, and are written once to specify how each record should be read out from any valid instantiation. As a result, a prediction is counted as graph-recoverable only if it can be reconstructed through these schema-derived queries. Figure~\ref{fig:fig1_core_results} summarises aggregate performance, class imbalance, and per-paper variability. Figure~\ref{fig:fig2_ablation_constraint} isolates the effects of external grounding services and constraint feedback. Figure~\ref{fig:fig3_error_anatomy} analyses dominant step-level error sources and highlights priorities. Exact scores and ablations appear in Tables~\ref{tab:overall} and \ref{tab:ablation}. Our evaluation is designed to assess interaction-based constraint enforcement by analysing its effects on graph recoverability and accuracy. 

\subsection{Ontology-compiled MCP tools enable semantically valid and content-correct end-to-end graph instantiation}
To assess whether ontology-compiled MCP tools produce outputs that are both semantically valid under the ontology and content-correct, we evaluate end-to-end graph-recoverable performance and summarise the results in Figure~\ref{fig:fig1_core_results}. Figure~\ref{fig:fig1_core_results}a shows strong overall performance (micro-F1 0.826; Table~\ref{tab:overall}), with high precision (0.844) and task-dependent recall (0.808). It also shows clear differences across categories. Synthesis steps perform best overall (F1 0.843). Reaction chemicals have perfect precision (1.000) but much lower recall (0.582). This suggests the system often avoids uncertain chemical mentions and misses some valid ones, especially when names are written in inconsistent forms (Table~\ref{tab:overall}).

These scores reflect both semantic and content correctness. A prediction counts as correct only if it is instantiated with the right entity types, relations, and required fields. It must also be recoverable by fixed SPARQL queries that reconstruct the expected JSON records. Errors in either content or structure, including missing individuals, wrong types, missing or incorrect links, or unfilled required slots, prevent recovery and are counted as false negatives.

Figure~\ref{fig:fig1_core_results}b quantifies the benchmark imbalance. This explains why micro-aggregates are dominated by high-volume categories and motivates reporting macro-averaged scores (macro precision 0.865, macro recall 0.735, macro F1 0.785; Table~\ref{tab:overall}). Figure~\ref{fig:fig1_core_results}c shows broad per-paper F1 distributions across all categories. Figure~\ref{fig:fig1_core_results}d indicates that recall variability drives the lower tail for some domains. Figure~\ref{fig:fig1_core_results} further shows that performance varies widely across papers. The best--worst contrasts suggest that differences in how authors report experiments, and how much concrete detail they include, have a strong impact on what the system can reliably extract and reconstruct from the graph.

In all, the results indicate that ontology-compiled MCP tools support semantically valid graph construction with strong content accuracy, while remaining errors are concentrated in recall-sensitive categories and in papers with sparse or unevenly reported evidence.

\subsection{Constraint feedback improves completeness of instantiated synthesis steps}

To assess the contribution and necessity of constraint feedback, we ablate the feedback channel and compare end-to-end instantiation scores. Figure~\ref{fig:fig2_ablation_constraint}a summarises the category-level impact of this ablation and shows that synthesis steps are among the most affected outputs. We therefore analyse synthesis-step behaviour in more detail.

In the ablated setting, we disable ontology-derived validation feedback at run time. We manually modify the AI-generated execution script and comment out the code paths that return feedback to the agent. This removes checks for required fields, unit and value normalization, step ordering and continuity, and other consistency constraints that are otherwise applied incrementally as synthesis steps are constructed. The agent therefore generates outputs without receiving signals about missing fields, invalid values, or structural violations.

Table~\ref{tab:ablation} shows a substantial drop in synthesis-step F1 when constraint feedback is removed, driven primarily by lower recall rather than precision. Figure~\ref{fig:fig2_ablation_constraint}b visualises this shift, showing that without feedback the system produces fewer synthesis-step instances that are complete enough to be recovered by the fixed SPARQL evaluation queries.

Figure~\ref{fig:fig2_ablation_constraint}c helps explain the underlying failure modes. Without feedback, step outputs more frequently violate basic structural expectations, including inconsistent or missing step numbers and unnecessary fragmentation into extra steps. The figure reports paired differences between full and no-feedback runs on the same papers using three proxy measures computed from the instantiated outputs. These capture increases in step-number inconsistencies, inflation in the number of steps, and regressions to placeholder or default values. Together, these results show that constraint feedback is important for guiding the model toward synthesis-step structures that are complete, well-formed, and reliably instantiable.


\subsection{Dominant synthesis-step error modes and improvement priorities}
To identify the main sources of remaining limitations in synthesis-step extraction and to guide future improvements, we analyse synthesis-step errors in detail using Figure~\ref{fig:fig3_error_anatomy}. Synthesis steps are the core output of the pipeline because they represent the ordered experimental procedure and link together materials, conditions, and outcomes. Errors at this level therefore have a direct impact on the usefulness of the extracted knowledge graph, which motivates a more detailed analysis than for the other categories.

Figure~\ref{fig:fig3_error_anatomy}a shows that synthesis-step errors are not evenly distributed across fields. A small number of fields account for a large share of the total errors. These include fields that encode step order, action descriptions, and key parameters. Errors arise both as missing fields, where required information is not extracted or instantiated, and as extra fields, where values are produced but do not correspond to the ground truth. This concentration indicates that overall performance is limited by a small set of recurring failure points rather than uniform noise across the schema.

Figure~\ref{fig:fig3_error_anatomy}b further separates fields by error type. Some fields are recall-limited, meaning the system often fails to extract values that are present in the paper. Other fields are precision-limited, meaning the system more often produces incorrect or unnecessary values. This distinction suggests different improvement strategies: recall-limited fields benefit from better coverage and normalization, while precision-limited fields require tighter constraints and filtering.

Building on this ranking, Figure~\ref{fig:fig3_error_anatomy}c presents an improvement roadmap under a conservative fix-by-field assumption. It estimates how synthesis-step F1 would increase if the highest-impact fields were corrected one at a time. The curve shows diminishing returns, where correcting only a few top contributors yields a large fraction of the achievable improvement.

Finally, Figure~\ref{fig:fig3_error_anatomy}d shows that synthesis-step errors are unevenly distributed across papers. A minority of papers accounts for a large fraction of the total errors, suggesting that targeted analysis of the most difficult papers can complement global field-level improvements.

\section{Discussion}\label{sec:discussion}


This work demonstrates a minimal constructive example of how generative models can be coupled to formal systems through executable semantics. Rather than treating symbolic knowledge as static context or validation targets, ontologies are operationalised as run-time constraints that shape agent behaviour through interaction. This shifts the role of large language models from text generators to stateful programs operating within a structured environment. 
More broadly, ontology-to-tools compilation suggests a pathway for operationalising symbolic constraints as run-time, feedback-driven mechanisms that shape model behaviour, rather than encoding them only as static representations.

\subsection{Limitations}\label{subsec:limitations}
The current evaluation covers one ontology and one document collection. We use a MOP-focused ontology and a benchmark of 30 MOP synthesis papers. The benchmark targets four information types: CBUs, characterisation entities, synthesis steps, and reaction chemicals. We did not test how results change when the ontology is updated and the pipeline is re-run. We also did not evaluate the approach in a different scientific domain.

Some extraction tasks remain harder than others. Chemical species identification shows high precision but lower recall, which means the system often misses valid mentions. This is likely due to variation in naming and reporting styles in the papers.

The main constraint on broader evaluation is the lack of curated datasets. Building reliable ground truth in new domains, or under revised ontologies, is still costly and time-consuming.



\subsection{Future work}\label{subsec:outlook}
Future work will directly address the limitations identified above by systematically evaluating robustness under ontology change. We will modify the ontology and re-run the full compilation and instantiation pipeline to quantify how executable tool interfaces and prompts adapt to evolving schemas, and to what extent regeneration can replace manual pipeline re-engineering. In parallel, we will apply the framework to additional scientific domains with distinct ontological structures to assess transfer beyond MOP synthesis.

Where suitable curated data is available, we will expand evaluation to larger and more heterogeneous corpora and to additional extraction targets. Particular emphasis will be placed on recall-limited tasks, especially chemical species identification and normalisation, while preserving the current level of precision, in order to better characterise the trade-offs introduced by enforcing ontological constraints at generation time. These experiments will allow us to quantify the stability of compiled action interfaces under schema and domain changes.

\section{Methods}\label{sec:methods}

The methodology is organized as a staged pipeline that separates ontology-driven compilation, ontology-constrained KG construction, and grounding.
The \emph{preparation stage} compiles an ontology T-Box and a small set of domain-agnostic meta-prompts into executable artefacts. These include a static JSON iteration plan, ontology- and task-specific instantiation prompts, and an ontology-specific MCP server that exposes ontology-aware construction and validation tools. The \emph{instantiation stage} executes the compiled plan for each document using a ReAct-style tool-use loop. It incrementally constructs A-Box individuals and relations in a persistent Turtle store and validates them as it goes. When constraints are violated, the tools return diagnostics that guide repair. The \emph{grounding stage} aligns selected constructed instance IRIs to canonical entities in an existing knowledge graph. It generates an ontology-conditioned lookup interface from the target T-Box and endpoint evidence. It then applies deterministic lexical matching to produce an explicit IRI mapping. This mapping is applied by rewriting instance IRIs or by adding explicit \texttt{owl:sameAs} links. Domain-specific derivation modules (\eg CBU derivation from crystallographic resources) are treated as downstream enrichment steps and are described separately from the core KG construction and grounding pipeline.

\subsection{Preparation stage}\label{sec:preparation}

The preparation stage converts an ontology T-Box into (i) a static, executable JSON plan for extraction and KG construction and (ii) an ontology-specific MCP server that exposes the operations referenced by that plan. The only manual inputs are a set of \emph{domain-agnostic meta-prompts for extraction and KG construction} that are reused across papers and domains. These prompts are provided in Listings~\ref{lst:system_prompt_iter_1}, \ref{lst:iter_1_meta_prompt}, \ref{lst:extension_system_prompt}, and \ref{lst:extension_user_prompt}. As summarised in Figures~\ref{fig:uml_mcp_creation} and~\ref{fig:iteration_example}, these meta-prompts guide an LLM to derive the task decomposition, synthesise ontology-aware
scripts, materialise an MCP server, and generate instantiation prompts capturing task-specific interpretation rules.

\paragraph{JSON-based task decomposition.}
A \emph{task decomposition agent} breaks the ontology-guided extraction problem into a small number of ordered steps. The result is written as a JSON plan, illustrated in Figure~\ref{fig:iteration_example}, which the instantiation agent can follow directly.

Each step in the plan states (i) what to do, (ii) what text prompt to use for extraction, and (iii) what prompt to use for knowledge-graph updates. The plan also lists what information each step reads and writes, such as the source paper, intermediate notes, and generated graph files. Optional sub-steps can be included for follow-up passes, such as enrichment or correction. Finally, the plan specifies which MCP tool groups are available at each step and which tools are needed, including their expected inputs and outputs. The outcome is a static, executable procedure that drives document processing in a fixed order.

\paragraph{Script and MCP tool generation from the T-Box.}
Given the T-Box and a set of domain-agnostic meta-prompts (Listings~\ref{lst:system_prompt_iter_1}, \ref{lst:iter_1_meta_prompt}, \ref{lst:extension_system_prompt}, and \ref{lst:extension_user_prompt}), the MCP Creation Agent generates an ontology-aware Python script that supports the classes and properties needed for the extraction scenario.

The script is built on top of a manually written, ontology-independent helper library. This library handles Turtle loading and saving, cross-step state management, deterministic IRI minting, and other generic utilities. The generated code is required to call these helpers rather than reimplement them. This keeps domain-specific logic separate from shared infrastructure.

The meta-prompts enforce a standard code structure. The script initialises an RDFLib graph, provides utilities to create and reuse instances by class, and defines functions to add links for each property. The generated code also distinguishes two kinds of constraints. \emph{Hard} constraints come from the formal T-Box axioms. They include class hierarchy, domain and range typing, datatype restrictions, and any modelled cardinalities. These axioms determine tool inputs and outputs and drive run-time checks that prevent invalid triples. \emph{Soft} constraints come from natural-language annotations in the ontology. For example, \texttt{rdfs:comment} may state inclusion or exclusion rules, naming conventions, or deduplication and reuse policies. These annotations are treated as guidance that complements, but does not override, the formal axioms.

\paragraph{MCP server construction.}
After the ontology-aware functions are generated, the MCP Integration Agent turns them into MCP tools. It reads each function signature and docstring and follows an integration meta-prompt. This process produces ontology-specific MCP servers, as shown in Figure~\ref{fig:uml_mcp_creation}.

The server exposes each function as an MCP tool with an ontology-derived name and a typed argument schema. It also attaches short usage instructions drawn from T-Box annotations. Tools are grouped into simple tool sets that match the JSON task plan, such as entity creation, attribute and relation completion, and cross-document linking. The resulting MCP server is then registered in the shared MCP tool pool and becomes available to all LLM-powered agents.

\paragraph{Design principles for instantiation MCP servers.}
Among the generated MCP servers, the instantiation MCP server is central at
runtime, and its design principles are enforced during preparation. First, tool
interfaces are defined in terms of ontology concepts, relations, and constraints:
the meta-prompts require the underlying scripts to respect which classes may be
connected by which properties, expected units, and (where applicable) reuse of
concepts from reference ontologies (e.g.\ OM for units). Second, tool
descriptions are ontology-guided: function signatures and natural-language
instructions are derived from the T-Box, including \texttt{rdfs:comment}
annotations that capture intended meanings and preferred usage patterns. Third,
the server is wired to a persistent Turtle (\texttt{.ttl}) store that acts as
cross-step memory, enabling reuse and incremental extension of previously
created instances across multiple iterations.

\paragraph{Instantiation prompt generation}\label{sec:prep_prompt_gen}
In addition to tools and scripts, the preparation agent generates
\emph{domain- and task-specific instantiation prompts} that capture \emph{soft
interpretation constraints} not reliably encoded as formal axioms. These prompts
encode operational definitions and heuristic decision rules that guide boundary
setting and classification during extraction (\eg what counts as a synthesis
step and how to delimit it; how to recognise a \texttt{HeatChill} step; how to
infer experimental atmosphere such as air vs.\ inert from textual cues). The
instantiation agent uses these prompts alongside the JSON plan and the MCP tool
schemas to decide what evidence to extract and when to invoke which tools; hard
schema constraints are enforced by the ontology-aware tools during instance
construction.
 
\subsection{Instantiation stage}

\begin{figure}
    \centering
    \includegraphics[width=0.95\linewidth]{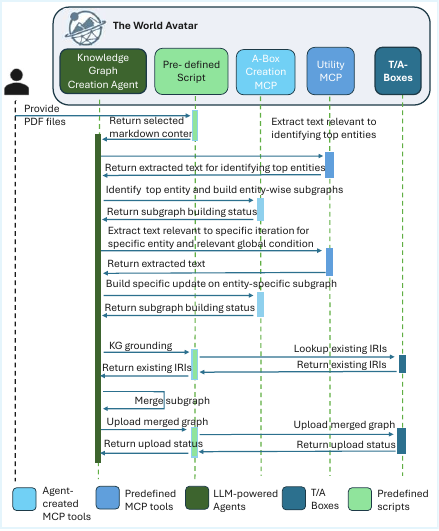}
   
    \caption{Ontology-to-tools compilation workflow. A preparation agent takes as input the
    ontology T-Box and a small set of manually authored, domain-agnostic meta-prompts
    for extraction and KG construction. It synthesises (i) an ontology-specific MCP
    server exposing ontology-aware tools with machine-checkable schemas and
    (ii) domain- and task-specific instantiation prompts (for synthesis steps,
    reaction chemicals, characterisation entities, and CBUs) that steer the runtime
    agent. A ReAct-style instantiation agent then executes the generated prompts and
    invokes the MCP tools to construct knowledge-graph instances from documents.}
    \label{fig:uml-kg-building}
\end{figure}

\begin{figure}[!htbp]
    \centering
    \includegraphics[width=0.93\linewidth]{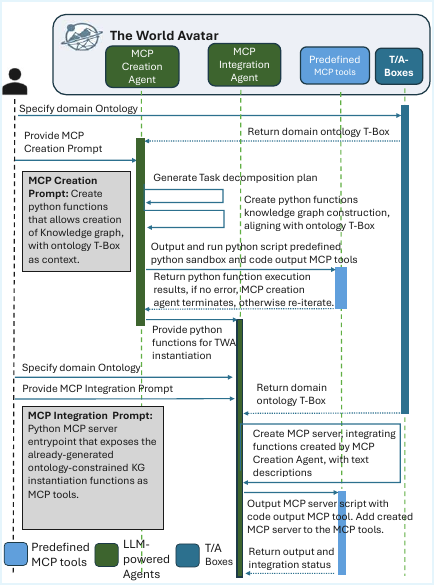}
    \caption{UML workflow of the two-agent framework for ontology-aligned knowledge-graph construction within the World Avatar (TWA) environment. The \textbf{MCP Creation Agent} runs in the TWA context and takes the domain ontology (T-Box; schema) and an MCP Creation Prompt to generate a KG instantiation library. The \textbf{MCP Integration Agent} then takes an MCP Integration Prompt plus the validated library (and its function descriptions) to package it into a deployable MCP server, exposing these functions as callable MCP tools and registering them into a shared MCP tool pool for reuse by downstream LLM agents, including TWA workflows.}
    \label{fig:uml_mcp_creation}
\end{figure}

\begin{figure}[!htbp] 
    \centering
    \includegraphics[width=1\linewidth]{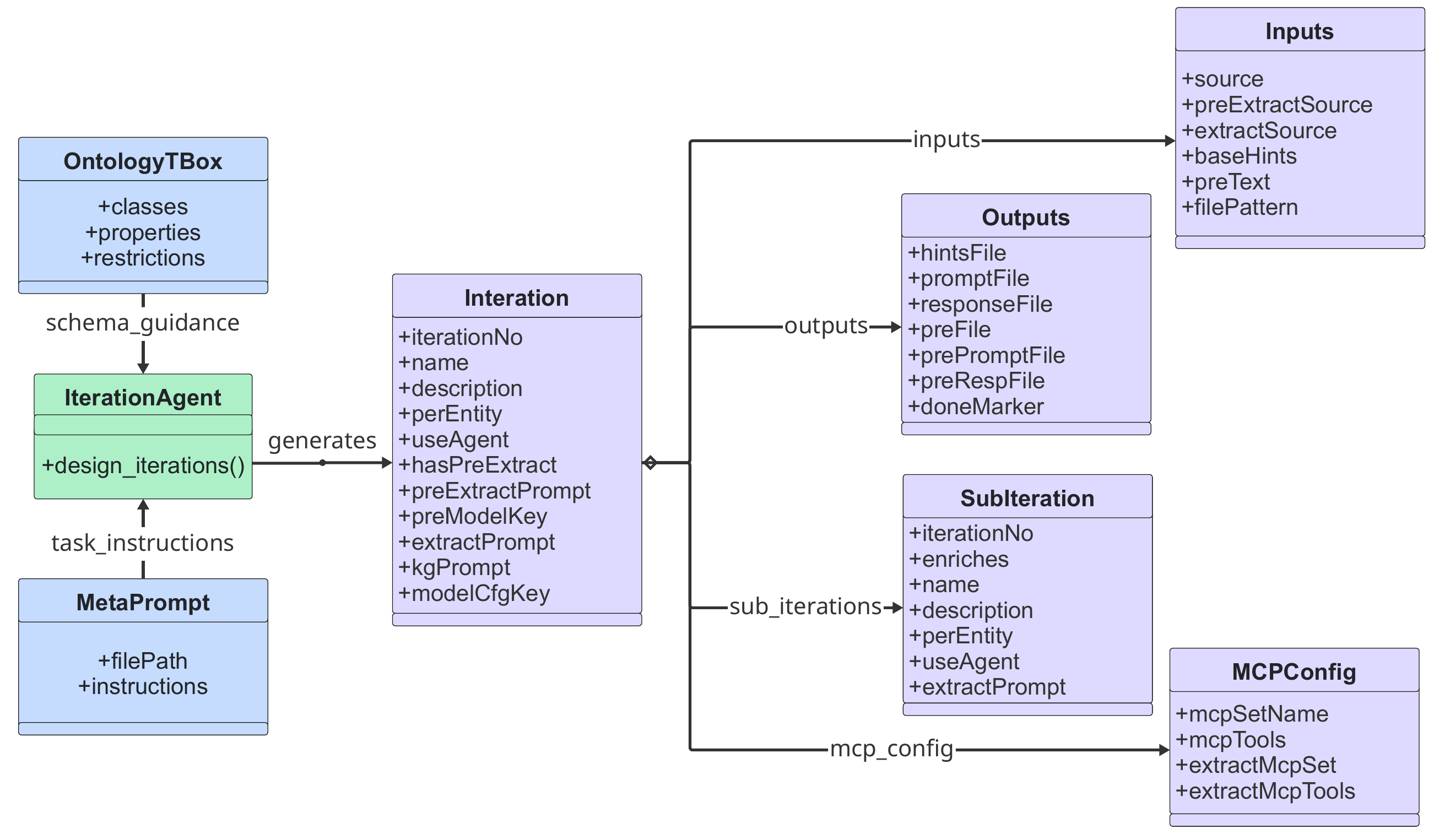}
    \caption{Class diagram of the structured task specification: a task decomposition model, guided by the Ontology T-Box and a meta-prompt, generates iterations that bundle 
    and KG-building steps, file flows (inputs/outputs), optional sub-iterations, and MCP tool configurations.}
    \label{fig:iteration_example}
\end{figure}

The instantiation stage executes the task specification produced during
preparation to construct ontology-aligned knowledge graphs from documents.
Guided by the JSON-based decomposition (Figure~\ref{fig:iteration_example}),
the runtime follows the iteration-oriented structure shown in
Figure~\ref{fig:uml-kg-building}, with an instantiation agent acting as a
ReAct-style controller over MCP tools and agent-generated, runtime intermediate
results (e.g.\ condensed passages and hints, extracted snippets, tool-call
inputs/outputs, logs, and completion markers), together with a persistent
Turtle store used for incremental KG updates.

\paragraph{Loading MCP tools for instantiation.}
Before processing a paper, the system loads two kinds of MCP servers. The first kind contains the ontology-specific instantiation tools generated in the preparation stage. The second kind provides shared utilities, such as external knowledge access and general text processing.

For chemical identifiers and properties, we use a third-party PubChem MCP server\footnote{\texttt{PubChem-MCP-Server} (GitHub): \url{https://github.com/JackKuo666/PubChem-MCP-Server}}. Access to crystallographic metadata is provided by a custom CCDC MCP server implemented in this work. A configuration file then lists the available tools and their input and output schemas. The instantiation agent reads this file so it can call both the local instantiation tools and the external PubChem and CCDC tools through a single, consistent interface.

\paragraph{Plan-driven ReAct execution.}
After the MCP tools are loaded, the instantiation agent follows the JSON plan step by step. Each step specifies the goal, the prompts to use, the files to read and write, and the tool groups that are allowed. Figure~\ref{fig:iteration_example} shows an example plan step.

The agent runs each step in a ReAct-style loop. It first reads the paper and produces intermediate notes, such as condensed passages, extracted snippets, normalised names, and lookup results. It then calls ontology-specific MCP tools to create entities, add attributes, and link relations in the Turtle store. A step ends when the expected outputs for that step have been produced, including any follow-up passes used for enrichment.

Each tool call returns both results and validation feedback. The feedback reports whether the requested update satisfies the ontology constraints. If a violation is detected, for example a missing required field, a type mismatch, or an invalid unit (Figure~\ref{fig:constraint_repair_example}), the tool returns an error with an explanation. The agent then retries with corrected inputs. It may also extract missing evidence from the paper or query external resources before calling the tool again.

\begin{figure}
  \centering
  \includegraphics[width=0.8\linewidth]{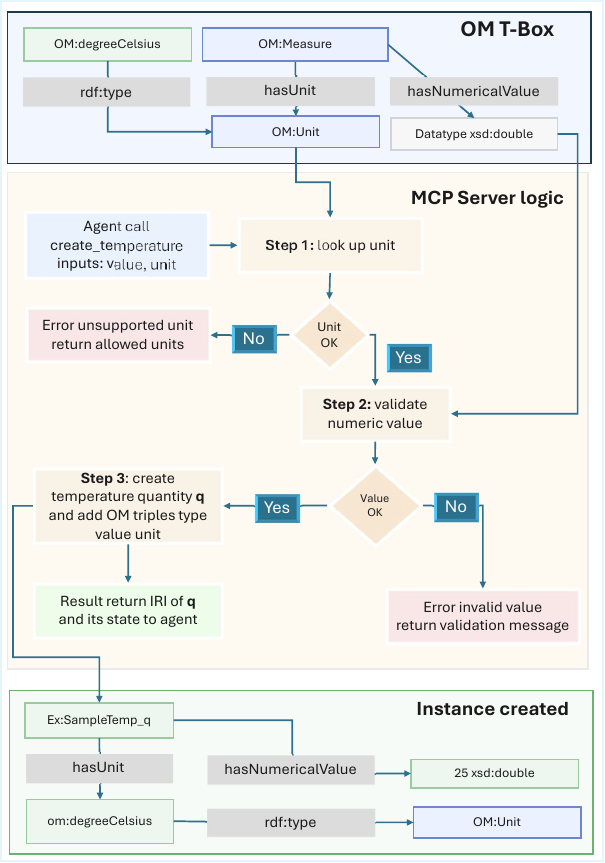}
  \caption{Illustrative example of the interaction between the OM T-Box
  (top), a single \texttt{create\_temperature} call within the MCP server
  logic (middle), and the resulting OM instance (bottom). The top box shows
  the ontology T-Box, which defines schema-level constraints for temperature
  quantities, such as the relevant classes and properties. The middle box
  shows an example script call that enforces these constraints by looking up
  the unit, validating the numeric value, and only then creating the
  corresponding OM individuals for the temperature value and its unit. The
  bottom box shows the resulting example instance that satisfies the T-Box
  constraints, linking the created quantity to a valid unit and a compliant
  numerical value.}
  \label{fig:ontology-code-instance}
\end{figure}

\paragraph{Ontology-constrained function calls.}
Ontology constraints are checked at tool-call time by the instantiation MCP server. The server functions follow the design rules set in the preparation stage. As illustrated in Figure~\ref{fig:ontology-code-instance}, a function such as \texttt{create\_temperature} looks up the relevant OM classes and properties. It checks the numeric value and unit against the T-Box. Only then does it create the corresponding individuals in the graph. Similar checks are applied to other quantities, relations, and class memberships.

Each call returns both the requested result and a check report. If the call violates a constraint, the server returns an error with an explanation. The instantiation agent uses this feedback in the next ReAct step to revise inputs, add missing fields, or trigger a repair action.

\paragraph{Turtle-based persistent memory.}
Throughout the instantiation stage, all MCP tools operate over a
Turtle-encoded knowledge graph (\texttt{.ttl}) that serves as persistent
cross-step memory. The instantiation MCP server reads from and writes to
this store, updating it after each successful creation, enrichment, or repair
operation. Instances created in earlier iterations can be looked up, linked,
and incrementally extended in later ones, and identifiers are reused according
to the deduplication and IRI-minting logic generated in the preparation
scripts. Intermediate Turtle snapshots, together with the file-level
\texttt{Inputs}/\texttt{Outputs} markers, record the state of the extraction
run at each iteration, so that processing can be resumed from a given point or
audited after the fact.

\subsection{Grounding stage}

After ontology-constrained KG construction, we apply a grounding stage to align constructed instance IRIs to canonical entities in a reference knowledge graph.
The purpose is to normalize identity across documents and pipelines by linking locally minted IRIs to existing KG IRIs, enabling deduplication and integration.
Grounding operates over selected ontology classes and produces a deterministic IRI mapping that is materialized either by rewriting IRIs or by adding \texttt{owl:sameAs} links.

\paragraph{Grounding tool generation}

Grounding tools are generated by LLM agents using the target ontology T-Box and KG endpoint evidence as the only domain-specific inputs.
The goal of this stage is to compile a grounding interface that supports canonicalization of constructed instances against a reference knowledge graph.

Given a target ontology schema (T-Box) and a SPARQL endpoint for the existing knowledge graph, the grounding generation agents automatically synthesize three tool scripts.
First, a sampling script analyzes the ontology schema and endpoint to identify relevant classes and label-like predicates and to estimate instance distributions.
Second, a label collection script collect and cache labels and identifiers for selected classes from the target KG and builds a local label index.
Third, a query and lookup interface script is generated that provides SPARQL accessors and deterministic fuzzy-lookup functions over the local label index.

The resulting query and lookup interface is exposed as an ontology-specific MCP server and becomes available as callable grounding tools in the runtime pipeline.
This generation process follows the same ontology-driven tool compilation paradigm as KG construction, while keeping grounding logic domain-agnostic at runtime.

\paragraph{Lexical grounding}

For each selected constructed instance, the grounding agent extracts one or more surface strings (\eg \texttt{rdfs:label} and ontology-specific alternative name properties) and queries the target-KG MCP lookup server, which returns a ranked, finite set of candidate matches from a locally cached label index.
The agent then applies a deterministic selection policy over this candidate set to choose a canonical target IRI and produces an explicit mapping from constructed IRIs to reference-KG IRIs.
Finally, the mapping is materialized either by rewriting IRIs in the RDF graph or by adding \texttt{owl:sameAs} links; the same procedure supports both single-file and batch grounding over collections of Turtle outputs.

\subsection{Preparation of evaluation dataset}\label{subsec:dataset}

We curated a benchmark of 30 scientific articles reporting MOP synthesis and
characterisation, focusing on papers where the target entities and relations are
explicitly described in the text. Ground truth was annotated in a predefined JSON
record format inherited from the prior (non-MCP) workflow and retained here to
enable like-for-like comparison across pipelines. This schema is aligned with the
ontology T-Box in the sense that each task category specifies the corresponding
entity types, relations, and attribute slots to be captured, but annotation is
performed directly in JSON rather than as ontology instances.

We manually populated the JSON records for
each paper following written guidelines that enforce consistent interpretation of
entity/slot definitions and alignment with T-Box intent (\eg, inclusion/exclusion
criteria and expected value types). The resulting benchmark contains 6{,}705
annotated items across synthesis steps, reaction chemicals, characterisation
entities, and chemical building units (CBUs) (Table~\ref{tab:benchmark_profile}).

\subsection{Evaluation metrics and methodology}

We evaluate end-to-end extraction \emph{coupled with ontology instantiation} via a
JSON-to-JSON protocol. For each task category, system outputs are first
instantiated as ontology individuals and relations in a Turtle graph. We then
apply a fixed, manually designed set of category-specific SPARQL queries (held
constant across all runs) to retrieve the target individuals, relations, and
literal attributes implied by the T-Box. Query bindings are deterministically
serialised into JSON records conforming to the same predefined schema used for
annotation. This evaluates instantiation quality rather than text extraction
alone: any missing individuals, missing links, or uninstantiated required slots
are not returned by the SPARQL queries and are therefore counted as false
negatives.

We compute precision, recall, and F1 by matching predicted and ground-truth JSON
records using \emph{slot-to-slot exact match} after basic normalisation.
Normalisation is limited to non-semantic formatting fixes (\eg, trimming
whitespace, normalising casing where appropriate, and applying simple canonical
forms for common unit strings when the schema expects a controlled label). For
record sets where ordering is not semantically meaningful (notably synthesis
steps), matching is \emph{order-insensitive}: predicted records are aligned to
ground truth under a one-to-one assignment that maximises the number of exactly
matched slots, and any unmatched predicted or ground-truth records contribute to
false positives or false negatives, respectively. Metrics are reported per
category and aggregated over all papers; we additionally report micro-aggregated
scores across all categories and macro-averaged scores across categories.

The same SPARQL-to-JSON conversion and matching procedure is applied to all
comparative baselines and ablation settings, ensuring that all results reflect
the same end-to-end criterion of producing evaluation-recoverable, correctly
instantiated outputs. Finally, CBU items correspond to \emph{grounded/derived}
CBUs constructed by combining paper evidence with database-backed normalisation
and derivation; they are therefore evaluated as an end-to-end grounding task
rather than a pure surface-form extraction task.

\subsection{CBU derivation for metal-organic polyhedra}

In addition to ontology-driven extraction from textual synthesis procedures, the framework instantiates an agent for automated Chemical Building Unit (CBU) derivation for metal--organic polyhedra (MOPs). This agent is implemented as a ReAct-style controller that orchestrates MCP-based access to crystallographic and chemical databases, and then materialises the resulting CBUs directly into The World Avatar knowledge graph.

The CBU derivation workflow proceeds as follows:

\begin{itemize}
  \item \textbf{MCP-based access to crystallographic data.} Given a MOP identifier (\eg  a CCDC deposition code), the agent uses an MCP server wrapping the CCDC database to locate and download the corresponding \texttt{.res} and \texttt{.cif} files. A Python-backed MCP tool parses these files to extract the asymmetric unit, symmetry operations, and atom/site information required for structural analysis.

  \item \textbf{Identification of CBUs.} Using the parsed crystallographic information, the agent invokes MCP tools for graph-based analysis of the coordination network (\eg  bond connectivity, coordination environments, linker topology). These tools segment the structure into metal nodes and organic linkers, and classify them into reusable CBU types suitable for downstream materials design workflows.

  \item \textbf{External chemical enrichment via PubChem and related services.} For each organic linker candidate, the agent calls MCP servers that wrap external databases such as PubChem to retrieve standard identifiers (InChI, InChIKey, SMILES), synonyms, and basic physicochemical properties. This enrichment step normalises the CBUs against widely used chemical identifiers and supports cross-database linkage.

  \item \textbf{Ontology-aligned CBU instantiation.} The derived and enriched CBUs are then passed to the instantiation MCP server, which creates ontology-consistent instances representing metal nodes, linkers, and their connectivity. The resulting MOP CBU descriptions are written into the Turtle-based persistent memory and become part of the shared knowledge graph that can be reused by subsequent extraction and reasoning tasks.
\end{itemize}

By combining MCP-based access to CCDC, PubChem, and ontology-aware instantiation tools, the agent autonomously bridges crystallographic data and ontology-level CBU representations for MOPs, without manual scripting or ad hoc intermediate formats.

\subsection{Models and inference settings}\label{sec:models}
Components in the pipeline are driven by large language models (LLMs), with different models assigned to different roles. Unless otherwise stated, we use deterministic decoding for tool-calling (\texttt{temperature}=0) and enforce schema validation on structured outputs at tool boundaries. For non-tool free-form generations (\eg narrative rationales or intermediate notes), we retain the same decoding settings unless explicitly noted to ensure reproducibility across runs.

\paragraph{Model assignments.}
We use \texttt{gpt-4.1} for document-level extraction; \texttt{gpt-4o} for knowledge-graph instantiation and construction actions; \texttt{gpt-5} for chemical building unit (CBU) derivation as well as script and prompt generation; and \texttt{gpt-4o-mini} for the agent-based query interface, where latency and cost are prioritised.

\paragraph{Inference and validation.}
Tool calls are executed with deterministic decoding (\texttt{temperature}=0) to minimise stochastic variation in argument selection and to make constraint-violation feedback comparable across runs. Structured outputs produced at tool boundaries are validated against the corresponding schemas, and violations are surfaced to the agent as explicit error messages to drive iterative repair until a valid instance is produced.

\paragraph{Token usage and cost.}
Script and prompt generation incurred a total LLM cost of \$5.70. For end-to-end processing of 30 papers, the total LLM cost for document-level extraction and instantiation/grounding actions was \$100.56 (\$71.61 + \$28.95), corresponding to an average per-article cost of \$3.35.


\section*{Data availability}
The curated benchmark dataset (30 MOP synthesis papers with $>$6{,}000 annotations), together with the curated ground truth, output files, and evaluation data used in this study, are available via the Cambridge Data Repository (\doi{10.17863/CAM.126228}) and via Dropbox\footnote{https://www.dropbox.com/scl/fi/u0dtyhyfa6cp7cr60jckq/Data-Public.zip}.

\section*{Code availability}
Code for the ontology-to-tools compilation workflow, including the MCP servers/tools and evaluation scripts, will be made available on GitHub\footnote{https://github.com/TheWorldAvatar/mcp-tool-layer/}.

\section*{Acknowledgements}
This research was supported by the National Research Foundation, Prime Minister’s Office, Singapore under its Campus for Research Excellence and Technological Enterprise (CREATE) programme. This project has received funding from the European Union’s Horizon Europe research and innovation programme under grants 101074004 (C2IMPRESS), 101188248 (CLIMATE-ADAPT4EOSC), and 101226137 (TOGETHER).

M.K.\ gratefully acknowledges the support of the Alexander von Humboldt Foundation and the Massachusetts Institute of Technology. S.D.R.\ acknowledges financial support from Fitzwilliam College, Cambridge, and the Cambridge Trust.

We thank all contributors who assisted with data collection, annotation, and quality control for the manually curated MOP synthesis benchmark. In particular, we thank Mingxi Lu, Yichen Sun, Yuan Gao, Kuhan Thayalan, and Matthew Olatunji for annotation support.

For the purpose of open access, the author has applied a Creative Commons Attribution (CC BY) licence to any Author Accepted Manuscript version arising from this submission.

\section*{Competing interests}
The authors declare no competing interests.

\section*{Author contributions}

X.Z.\ implemented the system, conducted the experiments, and wrote the manuscript. P.B.\ and C.Y.\ led data curation and annotation. T.A.\ contributed to data curation. J.A.\ contributed to manuscript writing. S.R.\ contributed to data curation and provided domain expertise. M.K.\ contributed to ideation, conceptualisation, manuscript writing, and funding acquisition. All authors reviewed and approved the final manuscript.

\clearpage

\section{Supplementary Information}\label{sec:supplementary}

\numberwithin{table}{section}
\numberwithin{figure}{section}
\numberwithin{equation}{section}

\if\sectionprefix\empty%
\else%
    \renewcommand{\thesection}{\sectionprefix}
\fi%

\subsection{Supplementary Note 1: Background and related work}\label{sec:supp-note-background}

\subsubsection{LLM-based agents and ReAct-style reasoning}
Large language model (LLM)--based agents couple a generative model with an explicit
action space, where each step conditions on a textual context (for example, a user
request, intermediate results, and tool outputs) and selects the next action such as
calling a tool, querying a knowledge graph, or producing a natural-language response.~\cite{yao2023react,schick_toolformer_2023,wu_autogen_2023,wang_survey_2024}
External capabilities are typically exposed through structured interfaces (for example,
function signatures with typed arguments), and tool responses are returned as additional
context to enable multi-step interaction with external systems.~\cite{schick_toolformer_2023,wang_survey_2024}

ReAct-style agents organise this interaction as an explicit loop of \emph{reasoning} and
\emph{acting}, alternating between intermediate reasoning traces and tool actions, and
incorporating tool observations into subsequent decisions.~\cite{yao2023react}
In information extraction, this framing supports decomposing long inputs, validating
intermediate structures, and iteratively refining partial outputs through repeated
reasoning--acting cycles.

\subsubsection{Tool-calling interfaces and the Model Context Protocol}
Earlier tool-calling interfaces for LLMs typically expose application-specific functions
through proprietary schemas or prompt templates, and integration logic is often implemented
separately for each deployment.~\cite{dave_chatgpt_2023,liu_we_2024,li_large_2024}
The Model Context Protocol (MCP) instead defines an open client--server protocol for
connecting LLM applications to external tools and data sources. In MCP, servers register
tools, resources, and prompts with machine-readable schemas; clients discover these
capabilities and request tool invocations through a standardised messaging layer.~\cite{mcp_spec_2025,openai_mcp_docs_2025}
MCP supports structured tool definitions, streaming of results, and standardised error
reporting across heterogeneous hosts and models, enabling reuse of the same tool servers
across different LLM applications.

\subsubsection{The World Avatar and its chemistry ontologies}
The World Avatar (TWA) is a cross-domain digital ecosystem that represents entities,
processes, and their interdependencies using ontologies and RDF knowledge graphs, accessed
and modified by software agents at runtime.~\cite{theworldavatar2025}
The reticular-chemistry stack of TWA is organised around two domain-generic core ontologies
and one application ontology. OntoSpecies represents chemical species, identifiers, and
characterisation data (with extensions for IR and elemental analysis). OntoSyn
(OntoSynthesis) represents synthesis procedures as sequences of \texttt{SynthesisStep}
operations transforming input \texttt{Species} into \texttt{ChemicalOutput} products.
OntoMOPs acts as an application ontology for metal--organic polyhedra (MOPs), modelling MOP
instances and their building units and linking OntoSyn procedures and OntoSpecies reactants
into the MOP design space.~\cite{theworldavatar2025,kondinski2022,pascazio2023,rihm2025}
Previous work implemented a MOP synthesis pipeline in TWA that used LLMs guided by
ontology-aligned JSON schemas to populate these chemistry knowledge graphs.~\cite{rihm2025}

\subsubsection{LLM-based knowledge instantiation and ontology grounding}
Large language models are increasingly used to construct and maintain knowledge graphs, by
inducing schemas, populating instances, or interacting with graph backends.~\citep{liang_survey_2024,ren2024survey}
From the perspective of this work, two lines are most relevant: schema-level approaches that
define or refine extraction targets, and instance-level approaches that extract and validate
concrete assertions.

\paragraph{Schema-level (T-Box).}
Schema-level pipelines treat the extraction schema as a contract that defines what types,
fields, and constraints outputs should satisfy. PARSE refines JSON Schemas as first-class
artefacts for extraction, but optimises primarily syntactic constraints encoded at the
JSON-schema level rather than ontology axioms.~\citep{shrimal2025parse}
LLMs4SchemaDiscovery mines candidate scientific schemas from text with human feedback, again
focusing on schema discovery rather than using a fixed ontology T-Box as the contract.~\citep{sadruddin2025llms4schema}
SCHEMA-MINERpro adds ontology grounding by mapping discovered schemas to reference ontologies,
but grounding remains a stage separate from instance creation.~\citep{sadruddin2025schemaminpro}
AutoSchemaKG induces schemas and triples at scale without a predefined domain T-Box, enabling
cross-domain graphs but decoupling extraction from ontology-encoded constraints such as units
and identification rules.~\citep{bai2025autoschemakg}

\paragraph{Instance-level (A-Box).}
Instance-level pipelines focus on extracting entities, relations, and events as concrete
assertions. KGGen uses strict JSON specifications to extract and refine triples, improving
structural consistency, but semantic validity with respect to a concrete ontology is enforced
via post-hoc checks rather than constraint-aware instance construction.~\citep{mo2025kggen}
Other pipelines align extracted relations to ontology predicates or use ontologies for
filtering and validation, but typically keep ontology enforcement separate from the tool
interfaces that govern how instances are created.~\citep{kommineni2024,olasunkanmi2025relate,khorshidi2025odkeplus}

\paragraph{MCP-assisted LLM-based solutions.}
MCP has emerged as a general mechanism for connecting LLM applications to tools and data
sources.~\citep{mcp_spec_2025,openai_mcp_docs_2025} Existing MCP-based
knowledge-graph servers often expose generic CRUD-style interfaces over graph backends,
demonstrating tool-based access but typically leaving ontological correctness to application
logic or post-hoc validation rather than compiling T-Box axioms into tool signatures and
runtime checks.~\citep{gannon2025mementomcp}

\paragraph{TWA agent composition framework}
Prior to recent LLM-based agent systems, The World Avatar (TWA) explored how semantic descriptions can support the \emph{discovery}, \emph{composition}, and \emph{execution} of software agents within a knowledge-graph ecosystem. ~\citep{zhou2019agent} introduced a semantic agent composition framework for TWA, in which agents are described using a lightweight agent ontology and grounded execution metadata to enable automated composition of agent workflows across domains.

This work motivates a view of ontologies not only as data models but also as interfaces to execution, where semantic descriptions govern which agents can be composed and how they may interact. In contrast to agent discovery and composition based on pre-defined interfaces, the present work compiles domain ontologies into executable tool interfaces and run-time constraints that directly regulate generative behaviour.
\subsection{Supplementary Methods}\label{sec:supp-methods}
This section provides extended implementation and evaluation details referenced in the main Methods, including dataset curation and annotation protocol, evaluation metrics and matching rules (including the CBU formula-only criterion), MCP server/tool specifications, and runtime policies (iterations, error handling, and repair strategies).

\clearpage
\subsubsection{Iteration 1: System Meta Prompt}

\begin{lstlisting}
You are a knowledge graph construction prompt expert specializing in ITERATION 1 KG building prompts.

ITERATION 1 is special: it only creates top-level entity instances from paper content, WITHOUT creating any related entities (inputs, outputs, sub-components, or detailed steps).

Your task:
1. Analyze the provided ontology (T-Box) to identify the top entity type (the main class that represents the overall procedure/process)
2. Extract all relevant rules, constraints, and identification guidelines from rdfs:comment annotations
3. Generate a focused, tool-oriented KG building prompt for ITERATION 1

CRITICAL CONSTRAINTS FOR ITER1:
- ONLY create instances of the top-level entity class (one per procedure described in the paper)
- Do NOT create any related entities (inputs, outputs, sub-components, steps) in this iteration
- Link each top-level entity to its source document
- Apply strict identification rules from the ontology to determine what qualifies as a valid instance
- Include all cardinality, scope, exclusion, and linking rules from the ontology

OUTPUT REQUIREMENTS:
- Start with "Follow these generic rules for any iteration." followed by the global rules
- Include MCP tool-specific guidance (error handling, IRI management, check_existing_* usage)
- Include explicit identification section (document identifier handling)
- Clearly state the task scope: create top-level entities only, no related entities
- List all constraints from the ontology rdfs:comment for the top-level entity class
- Include clear termination conditions
- Be concise and tool-oriented (this prompt is for an MCP agent using function calls)
- Be completely domain-agnostic (no specific compound types, no domain-specific terminology)

Do NOT include:
- Domain-specific examples (\eg specific compound names, specific synthesis types)
- Variable placeholders like {doi} or {paper_content} - these will be added programmatically
- Verbose ontology explanations - focus on actionable rules
- Any mention of specific chemical entities, materials, or domain-specific concepts
\end{lstlisting}\label{lst:system_prompt_iter_1}

\subsubsection{Iteration 1: User Meta Prompt}

\begin{lstlisting}
Based on the ontology and available MCP tools below, generate a KG building prompt for ITERATION 1.

ITERATION 1 SCOPE:
- Create ONLY top-level entity instances (the main procedure/process class)
- Do NOT create any related entities (inputs, outputs, sub-components, steps)
- Link to source document

ONTOLOGY (T-Box):
{tbox}

MCP MAIN SCRIPT (Available Tools) [Python]:
{mcp_main_script}

The MCP Main Script shows all available tools with their descriptions, parameters, and usage guidance. Use this to understand what tools are available and how they should be called.

REQUIREMENTS:
1. Extract ALL rules from the top-level entity class rdfs:comment, especially:
   - Scope (what qualifies as a valid instance)
   - Different forms / methods / variations rules
   - Exclusions (what NOT to create)
   - Cardinality requirements
   - Linking requirements
   - Conservative behavior guidelines
   - Critical exclusions for extraction

2. Include global MCP rules:
   - Tool invocation rules (never call same tool twice with identical args)
   - IRI management (must create before passing, use check_existing_* tools)
   - Error handling (status codes, already_attached, retryable)
   - Placeholder policies
   - Termination conditions (run_status: done)

3. Include identification section:
   - Document identifier handling (treat as sole task identifier, reuse consistently)
   - Entity focus guidance (for when entity_label/entity_uri provided)

4. Be concise and actionable - this is for an MCP agent making function calls

5. Be completely domain-agnostic:
   - Do NOT mention specific compound types, materials, or chemical entities
   - Use generic terminology that applies to any domain
   - Adapt wording from the ontology to be domain-neutral where possible

Generate the prompt now (do NOT include variable placeholders - those will be added programmatically):
\end{lstlisting}\label{lst:iter_1_meta_prompt}

\subsubsection{Extension Ontology: System Meta Prompt}

\begin{lstlisting}
You are an expert in creating knowledge graph building prompts for extension ontologies.

Extension ontologies are simpler ontologies that extend a main ontology with additional specialized information. They use MCP tools to build A-Boxes that link to the main ontology's A-Box.

Your task is to analyze a T-Box ontology and MCP tools to create a KG building prompt that:
1. Provides a clear task route for building the extension A-Box
2. Emphasizes using MCP tools to populate the A-Box
3. Requires comprehensive population (making certain MCP function calls compulsory)
4. Emphasizes IRI reuse from the main ontology A-Box
5. Includes domain-specific requirements extracted from the T-Box comments
6. Forbids fabrication - only use information from the paper

CRITICAL RULES:
- Read ALL classes, properties, and rdfs:comment fields in the T-Box
- Extract domain-specific requirements from rdfs:comment (\eg required fields, cardinality constraints)
- Focus on HOW to build the A-Box, not just WHAT to extract
- Emphasize the connection between the extension A-Box and the main ontology A-Box
- Make the prompt actionable with clear steps
- Output ONLY the prompt text (no markdown fences, no commentary)
\end{lstlisting}\label{lst:extension_system_prompt}

\subsubsection{Extension Ontology: User Meta Prompt}

\begin{lstlisting}
Generate a KG building prompt for an extension ontology.

T-Box (analyze to understand the ontology structure and requirements):
```turtle
{tbox}
MCP Main Script (understand available tools, their parameters, and calling sequences):

python
 
{mcp_main_script}
Your prompt MUST:

State the task - Extend the main ontology A-Box with the extension A-Box

Provide a task route - Give a recommended sequence of steps for building the KG

Emphasize MCP tools - Instruct to use the MCP server to populate the A-Box

Require IRI reuse - Emphasize reusing existing IRIs from the main ontology A-Box

Extract T-Box requirements - Include any compulsory requirements from rdfs:comment (\eg required fields, minimum instances)

Forbid fabrication - Only use information from the paper content

Include tool-specific notes - If the T-Box or domain requires special handling (\eg external database integration, data transformations), include those notes

Structure your output as:

css
 
Your task is to extend the provided A-Box of [MainOntology] with the [extension ontology] A-Box, according to the paper content.

You should use the provided MCP server to populate the [extension ontology] A-Box.

Here is the recommended route of task:

[Step-by-step guidance based on T-Box structure]

Requirements:

[List of requirements based on T-Box rdfs:comment and MCP tool constraints]

Special note:

[Any domain-specific notes based on T-Box comments]

Here is the DOI for this run (normalized and pipeline forms):

- DOI: {{doi_slash}}
- Pipeline DOI: {{doi_underscore}}

Here is the [MainOntology] A-Box:

{{main_ontology_a_box}}

Here is the paper content:

{{paper_content}}

CRITICAL:

Extract domain-specific requirements from the T-Box rdfs:comment fields. Do NOT invent requirements. ALL requirements must be justified by the T-Box or MCP tool constraints.

Output EXACTLY the structure shown above. Do NOT add any additional sections after {{paper_content}}. This is the END of the prompt.

Generate the prompt now:
\end{lstlisting}\label{lst:extension_user_prompt}

\clearpage
\subsubsection{Illustrative trace of constraint-triggered repair}

Listing~\ref{fig:constraint_repair_example} shows a representative repair cycle
triggered by an ontology constraint on measurement units. The ontology
restricts temperature units to a controlled vocabulary that includes
\emph{degree Celsius}. When the instantiation agent first instantiates a
temperature condition using the verbatim unit token \emph{C} from text, the
ontology-aware MCP tool rejects the input and returns a structured validation
error listing allowed values. The agent then normalises the unit to an allowed
entry and retries, producing a valid instance.

\begin{lstlisting}[caption={Representative MCP trace: ontology-constrained unit repair (C $\rightarrow$ degree Celsius).},label={fig:constraint_repair_example}]
Input evidence (paper text):
  "... heated at 120 C for 12 h ..."

Attempt 1 (verbatim unit token):
  Tool: mop.create_temperature_condition
  Input:    { "context": "reaction_step_3",
              "value": 120,
              "unit": "C" }

  Tool response (constraint violation):
    { "ok": false,
      "error_type": "OntologyConstraintViolation",
      "field": "unit",
      "message": "Unit value 'C' is not permitted by the ontology.",
      "allowed_values": ["degree Celsius", "kelvin"]
    }

Repair (from tool feedback):
  - Map shorthand unit token "C" to the allowed vocabulary entry "degree Celsius".
  - Retry tool call with corrected unit.

Attempt 2 (normalised unit):
  Tool: mop.create_temperature_condition
  Input:
    { "context": "reaction_step_3",
      "value": 120,
      "unit": "degree Celsius"
    }

  Tool response (success):
    {
      "ok": true,
      "instance_iri": "twa:TemperatureCondition_0f3a...",
      "validated": true
    }
\end{lstlisting}

\subsection{Supplementary Evaluation Results}

Tables~\ref{tab:steps_full}--\ref{tab:char_full} report per-paper extraction performance against the full ground truth for the three evaluated outputs: synthesis steps (Table~\ref{tab:steps_full}), chemical building units under a formula-only matching criterion (Table~\ref{tab:cbu_formula_only}), and characterisation entities (Table~\ref{tab:char_full}). For each paper (indexed by DOI), we provide true positives (TP), false positives (FP), false negatives (FN), and the derived precision, recall, and F1 scores, followed by an overall aggregate row.

\begin{table}[!hpbt]
\centering
\caption{Benchmark profile (30 papers). Ground-truth positives are computed as TP$+$FN from Table~\ref{tab:overall}.}
\label{tab:benchmark_profile}
\begin{tabular}{lrr}
\hline
Category & Papers & Ground-truth positives \\
\hline
CBU              & 30 & 164 \\
Characterisation & 30 & 926 \\
Steps            & 30 & 4940 \\
Chemicals        & 30 & 675 \\
\hline
Total            & 30 & 6705 \\
\hline
\end{tabular}
\end{table}

\begin{table*}[t]
\centering
\caption{Overall evaluation summary by category (system output recovered from instantiated graphs vs.\ full ground truth). CBU denotes grounded/derived CBUs (Methods).}
\label{tab:overall}
\begin{tabular}{lrrrrrr}
\hline
Category & TP & FP & FN & Precision & Recall & F1 \\
\hline
CBU              & 128  &  40 &  36 & 0.762 & 0.780 & 0.771 \\
Characterisation & 669  & 102 & 257 & 0.868 & 0.722 & 0.788 \\
Steps            & 4225 & 857 & 715 & 0.831 & 0.855 & 0.843 \\
Chemicals        & 393  &   0 & 282 & 1.000 & 0.582 & 0.736 \\
\hline
\textbf{Micro-aggregate} & \textbf{5415} & \textbf{999} & \textbf{1290} & \textbf{0.844} & \textbf{0.808} & \textbf{0.826} \\
\textbf{Macro-average}   & --- & --- & --- & \textbf{0.865} & \textbf{0.735} & \textbf{0.785} \\
\hline
\end{tabular}
\end{table*}







\begin{table}[hpbt]
\centering
\caption{Ablation study. CBU denotes grounded/derived CBUs (Methods).}
\label{tab:ablation}
\begin{tabular}{lrrrr}
\hline
Setting & CBU F1 & Char. F1 & Steps F1 & Chem. F1 \\
\hline
Full system & \textbf{0.771} & \textbf{0.788} & \textbf{0.843} & \textbf{0.736} \\
-- external tools & 0.000 & 0.610 & 0.800 & 0.732 \\
-- constraint feedback & 0.768 & 0.788 & 0.572 & 0.724 \\
\hline
\end{tabular}
\end{table}

\begin{table*}[t]
\centering
\caption{Per-paper extraction results for synthesis steps against the full ground truth.}
\label{tab:steps_full}
\resizebox{\linewidth}{!}{%
\begin{tabular}{lrrrrrr}
\hline
DOI & TP & FP & FN & Precision & Recall & F1 \\
\hline
10.1002.anie.201811027~\cite{gong_bottomup_2019}              & 185 & 31 & 19 & 0.856 & 0.907 & 0.881 \\
10.1002.anie.202010824~\cite{gong_facedirected_2020}          & 305 & 10 & 19 & 0.968 & 0.941 & 0.955 \\
10.1002.chem.201604264~\cite{liu_controlled_2016}             & 245 & 28 & 30 & 0.897 & 0.891 & 0.894 \\
10.1002.chem.201700798~\cite{ju_coordination_2017}            &  47 & 10 &  8 & 0.825 & 0.855 & 0.839 \\
10.1002.chem.201700848~\cite{garai_selfexfoliated_2017}       & 111 & 20 & 30 & 0.847 & 0.787 & 0.816 \\
10.1021.acs.cgd.6b00306~\cite{rathnayake_investigating_2016}   &  35 & 10 &  4 & 0.778 & 0.897 & 0.833 \\
10.1021.acs.chemmater.8b01667~\cite{barreda_mechanochemical_2018} & 105 & 35 & 24 & 0.750 & 0.814 & 0.781 \\
10.1021.acs.inorgchem.4c02394~\cite{wang_steerable_2024}      & 216 & 38 & 44 & 0.850 & 0.831 & 0.840 \\
10.1021.acs.inorgchem.8b01130~\cite{gosselin_design_2018}     & 207 & 74 & 70 & 0.737 & 0.747 & 0.742 \\
10.1021.acsami.7b09339~\cite{park_chromiumii_2017}            & 117 & 35 & 40 & 0.770 & 0.745 & 0.757 \\
10.1021.acsami.7b18836~\cite{lee_porous_2018}                 &  80 & 24 &  6 & 0.769 & 0.930 & 0.842 \\
10.1021.acsami.8b02015~\cite{barreda_ligand-based_2018}       & 118 & 28 & 37 & 0.808 & 0.761 & 0.784 \\
10.1021.cg4018322~\cite{paul_secondary_2014}                   &  50 &  4 & 10 & 0.926 & 0.833 & 0.877 \\
10.1021.ic050460z~\cite{ke_synthesis_2005}                     & 139 & 63 & 41 & 0.688 & 0.772 & 0.728 \\
10.1021.ic402428m~\cite{liu_situ_2013}                         & 180 & 51 & 36 & 0.779 & 0.833 & 0.805 \\
10.1021.ic501012e~\cite{tan_two_2014}                          & 147 &  7 &  3 & 0.955 & 0.980 & 0.967 \\
10.1021.ic802382p~\cite{prakash_metalorganic_2009}             &  84 & 14 &  5 & 0.857 & 0.944 & 0.898 \\
10.1021.ja042802q~\cite{sudik_design_2005}                     & 389 &135 & 80 & 0.742 & 0.829 & 0.783 \\
10.1021.ja105986b~\cite{zheng_cubic_2010}                      &  63 & 23 & 25 & 0.733 & 0.716 & 0.724 \\
10.1021.jacs.7b00037~\cite{zhang_controlled_2017}              & 128 &  4 &  7 & 0.970 & 0.948 & 0.959 \\
10.1021.jacs.8b10866~\cite{zhang_self-assembly_2018}           & 214 & 51 & 28 & 0.808 & 0.884 & 0.844 \\
10.1039.C2CC34265K~\cite{du_giant_2012}                        & 166 & 22 & 33 & 0.883 & 0.834 & 0.858 \\
10.1039.C5CC05913E~\cite{augustyniak_vanadium_2015}            &  55 & 13 &  4 & 0.809 & 0.932 & 0.866 \\
10.1039.C5DT04764A~\cite{zhang_polyoxovanadate-based_2016}     & 161 &  0 &  3 & 1.000 & 0.982 & 0.991 \\
10.1039.C5RA26357C~\cite{chen_synthesis_2016}                  &  50 & 14 & 14 & 0.781 & 0.781 & 0.781 \\
10.1039.C6CC04583A~\cite{zhang_anderson-like_2016}             & 206 & 64 & 52 & 0.763 & 0.798 & 0.780 \\
10.1039.C6DT02764D~\cite{zhang_synthesis_2016}                 & 103 & 15 & 15 & 0.873 & 0.873 & 0.873 \\
10.1039.C7CC01208J~\cite{zhang_self-assembly_2017}             &  92 & 10 & 10 & 0.902 & 0.902 & 0.902 \\
10.1039.C8DT02580K~\cite{gong_functionalized_2018}             & 119 &  5 &  8 & 0.960 & 0.937 & 0.948 \\
10.1039.D3QI01501G~\cite{yang_efficient_2023}                  & 106 & 20 & 12 & 0.841 & 0.898 & 0.869 \\
\hline
Overall                                                         & 4223 & 858 & 717 & 0.831 & 0.855 & 0.843 \\
\hline
\end{tabular}
}
\end{table*}

\begin{table*}[t]
\centering
\caption{Per-paper formula-only extraction results for chemical building units (CBUs) against the full ground truth.}
\label{tab:cbu_formula_only}
\begin{tabular}{lrrrrrr}
\hline
DOI & TP & FP & FN & Precision & Recall & F1 \\
\hline
10.1021.acsami.7b18836~\cite{lee_porous_2018}                 &  3 &  1 &  1 & 0.750 & 0.750 & 0.750 \\
10.1039.C8DT02580K~\cite{gong_functionalized_2018}            &  4 &  0 &  0 & 1.000 & 1.000 & 1.000 \\
10.1002.chem.201700848~\cite{garai_selfexfoliated_2017}       &  4 &  2 &  2 & 0.667 & 0.667 & 0.667 \\
10.1021.acs.inorgchem.4c02394~\cite{wang_steerable_2024}      &  8 &  2 &  2 & 0.800 & 0.800 & 0.800 \\
10.1021.ic402428m~\cite{liu_situ_2013}                        &  8 &  0 &  0 & 1.000 & 1.000 & 1.000 \\
10.1039.C5RA26357C~\cite{chen_synthesis_2016}                 &  2 &  0 &  0 & 1.000 & 1.000 & 1.000 \\
10.1039.C6DT02764D~\cite{zhang_synthesis_2016}                &  2 &  2 &  2 & 0.500 & 0.500 & 0.500 \\
10.1021.ic501012e~\cite{tan_two_2014}                         &  4 &  0 &  0 & 1.000 & 1.000 & 1.000 \\
10.1002.chem.201604264~\cite{liu_controlled_2016}             &  8 &  2 &  2 & 0.800 & 0.800 & 0.800 \\
10.1021.jacs.7b00037~\cite{zhang_controlled_2017}             &  1 &  3 &  3 & 0.250 & 0.250 & 0.250 \\
10.1039.C7CC01208J~\cite{zhang_self-assembly_2017}            &  2 &  2 &  2 & 0.500 & 0.500 & 0.500 \\
10.1021.acschemmater.8b01667~\cite{barreda_mechanochemical_2018} &  6 &  2 &  0 & 0.750 & 1.000 & 0.857 \\
10.1039.C5DT04764A~\cite{zhang_polyoxovanadate-based_2016}    &  3 &  3 &  3 & 0.500 & 0.500 & 0.500 \\
10.1039.C6CC04583A~\cite{zhang_anderson-like_2016}            & 10 &  0 &  0 & 1.000 & 1.000 & 1.000 \\
10.1021.ic802382p~\cite{prakash_metalorganic_2009}            &  2 &  0 &  0 & 1.000 & 1.000 & 1.000 \\
10.1002.anie.202010824~\cite{gong_facedirected_2020}          &  7 &  3 &  3 & 0.700 & 0.700 & 0.700 \\
10.1021.acs.inorgchem.8b01130~\cite{gosselin_design_2018}     &  4 &  2 &  2 & 0.667 & 0.667 & 0.667 \\
10.1039.C2CC34265K~\cite{du_giant_2012}                       &  3 &  3 &  3 & 0.500 & 0.500 & 0.500 \\
10.1021.acsami.7b09339~\cite{park_chromiumii_2017}            &  8 &  0 &  0 & 1.000 & 1.000 & 1.000 \\
10.1002.anie.201811027~\cite{gong_bottomup_2019}              &  4 &  4 &  0 & 0.500 & 1.000 & 0.667 \\
10.1021.acs.cgd.6b00306~\cite{rathnayake_investigating_2016}  &  0 &  2 &  2 & 0.000 & 0.000 & 0.000 \\
10.1039.D3QI01501G~\cite{yang_efficient_2023}                 &  2 &  0 &  2 & 1.000 & 0.500 & 0.667 \\
10.1021.cg4018322~\cite{paul_secondary_2014}                  &  2 &  0 &  0 & 1.000 & 1.000 & 1.000 \\
10.1039.C5CC05913E~\cite{augustyniak_vanadium_2015}           &  2 &  0 &  0 & 1.000 & 1.000 & 1.000 \\
10.1021.ja105986b~\cite{zheng_cubic_2010}                     &  2 &  0 &  0 & 1.000 & 1.000 & 1.000 \\
10.1021.ic050460z~\cite{ke_synthesis_2005}                    &  6 &  0 &  0 & 1.000 & 1.000 & 1.000 \\
10.1021.jacs.8b10866~\cite{zhang_self-assembly_2018}          &  6 &  2 &  2 & 0.750 & 0.750 & 0.750 \\
10.1021.ja042802q~\cite{sudik_design_2005}                    &  7 &  5 &  3 & 0.583 & 0.700 & 0.636 \\
10.1021.acsami.8b02015~\cite{barreda_ligand-based_2018}       &  6 &  0 &  2 & 1.000 & 0.750 & 0.857 \\
10.1002.chem.201700798~\cite{ju_coordination_2017}            &  2 &  0 &  0 & 1.000 & 1.000 & 1.000 \\
\hline
Overall                                                        & 128 & 40 & 36 & 0.762 & 0.780 & 0.771 \\
\hline
\end{tabular}
\end{table*}
\begin{table*}[t]
\centering
\caption{Per-paper extraction results for characterisation entities against the full ground truth.}
\label{tab:char_full}
\resizebox{\linewidth}{!}{%
\begin{tabular}{lrrrrrr}
\hline
DOI & TP & FP & FN & Precision & Recall & F1 \\
\hline
10.1002.anie.201811027~\cite{gong_bottomup_2019}              & 26 &  0 &  4 & 1.000 & 0.867 & 0.929 \\
10.1002.anie.202010824~\cite{gong_facedirected_2020}          & 36 & 11 & 16 & 0.766 & 0.692 & 0.727 \\
10.1002.chem.201604264~\cite{liu_controlled_2016}             & 10 &  7 & 16 & 0.588 & 0.385 & 0.465 \\
10.1002.chem.201700798~\cite{ju_coordination_2017}            &  5 &  2 &  5 & 0.714 & 0.500 & 0.588 \\
10.1002.chem.201700848~\cite{garai_selfexfoliated_2017}       & 21 &  4 & 15 & 0.840 & 0.583 & 0.689 \\
10.1021.acs.cgd.6b00306~\cite{rathnayake_investigating_2016}   &  8 &  1 &  1 & 0.889 & 0.889 & 0.889 \\
10.1021.acs.chemmater.8b01667~\cite{barreda_mechanochemical_2018} & 36 &  3 & 10 & 0.923 & 0.783 & 0.847 \\
10.1021.acs.inorgchem.4c02394~\cite{wang_steerable_2024}      & 43 &  9 & 16 & 0.827 & 0.729 & 0.775 \\
10.1021.acs.inorgchem.8b01130~\cite{gosselin_design_2018}     & 38 &  1 & 11 & 0.974 & 0.776 & 0.864 \\
10.1021.acsami.7b09339~\cite{park_chromiumii_2017}            & 29 &  1 & 11 & 0.967 & 0.725 & 0.829 \\
10.1021.acsami.7b18836~\cite{lee_porous_2018}                 & 17 &  2 &  2 & 0.895 & 0.895 & 0.895 \\
10.1021.acsami.8b02015~\cite{barreda_ligand-based_2018}       & 19 &  2 &  8 & 0.905 & 0.704 & 0.792 \\
10.1021.cg4018322~\cite{paul_secondary_2014}                   & 13 &  1 & 17 & 0.929 & 0.433 & 0.591 \\
10.1021.ic050460z~\cite{ke_synthesis_2005}                     & 31 &  1 &  7 & 0.969 & 0.816 & 0.886 \\
10.1021.ic402428m~\cite{liu_situ_2013}                         & 20 &  0 & 24 & 1.000 & 0.455 & 0.625 \\
10.1021.ic501012e~\cite{tan_two_2014}                          & 14 &  4 &  6 & 0.778 & 0.700 & 0.737 \\
10.1021.ic802382p~\cite{prakash_metalorganic_2009}             &  7 &  2 &  5 & 0.778 & 0.583 & 0.667 \\
10.1021.ja042802q~\cite{sudik_design_2005}                     & 51 &  2 &  6 & 0.962 & 0.895 & 0.927 \\
10.1021.ja105986b~\cite{zheng_cubic_2010}                      &  8 &  2 &  5 & 0.800 & 0.615 & 0.696 \\
10.1021.jacs.7b00037~\cite{zhang_controlled_2017}              & 16 &  6 &  4 & 0.727 & 0.800 & 0.762 \\
10.1021.jacs.8b10866~\cite{zhang_self-assembly_2018}           & 41 &  8 &  9 & 0.837 & 0.820 & 0.828 \\
10.1039.C2CC34265K~\cite{du_giant_2012}                        & 23 &  8 & 13 & 0.742 & 0.639 & 0.687 \\
10.1039.C5CC05913E~\cite{augustyniak_vanadium_2015}            &  7 &  2 &  4 & 0.778 & 0.636 & 0.700 \\
10.1039.C5DT04764A~\cite{zhang_polyoxovanadate-based_2016}     & 24 &  2 &  6 & 0.923 & 0.800 & 0.857 \\
10.1039.C5RA26357C~\cite{chen_synthesis_2016}                  &  8 &  2 &  5 & 0.800 & 0.615 & 0.696 \\
10.1039.C6CC04583A~\cite{zhang_anderson-like_2016}             & 43 & 10 & 15 & 0.811 & 0.741 & 0.775 \\
10.1039.C6DT02764D~\cite{zhang_synthesis_2016}                 & 18 &  2 &  4 & 0.900 & 0.818 & 0.857 \\
10.1039.C7CC01208J~\cite{zhang_self-assembly_2017}             & 18 &  2 &  4 & 0.900 & 0.818 & 0.857 \\
10.1039.C8DT02580K~\cite{gong_functionalized_2018}             & 16 &  4 &  4 & 0.800 & 0.800 & 0.800 \\
10.1039.D3QI01501G~\cite{yang_efficient_2023}                  & 22 &  4 & 11 & 0.846 & 0.667 & 0.746 \\
\hline
Overall                                                         & 668 & 105 & 264 & 0.864 & 0.717 & 0.784 \\
\hline
\end{tabular}
}
\end{table*}

\begin{table*}[t]
\centering
\caption{Per-paper extraction results for chemicals against the full ground truth.}
\label{tab:chems_full}
\resizebox{\linewidth}{!}{
\begin{tabular}{lrrrrrr}
\hline
DOI & TP & FP & FN & Precision & Recall & F1 \\
\hline
10.1002.anie.201811027~\cite{gong_bottomup_2019}              &  8 &  0 &  6 & 1.000 & 0.571 & 0.727 \\
10.1002.anie.202010824~\cite{gong_facedirected_2020}          & 24 &  0 &  6 & 1.000 & 0.800 & 0.889 \\
10.1002.chem.201604264~\cite{liu_controlled_2016}             & 22 &  0 & 13 & 1.000 & 0.629 & 0.772 \\
10.1002.chem.201700798~\cite{ju_coordination_2017}            &  3 &  0 &  6 & 1.000 & 0.333 & 0.500 \\
10.1002.chem.201700848~\cite{garai_selfexfoliated_2017}       & 12 &  0 &  7 & 1.000 & 0.632 & 0.774 \\
10.1021.acs.cgd.6b00306~\cite{rathnayake_investigating_2016}   &  5 &  0 &  3 & 1.000 & 0.625 & 0.769 \\
10.1021.acs.chemmater.8b01667~\cite{barreda_mechanochemical_2018} &  7 &  0 &  1 & 1.000 & 0.875 & 0.933 \\
10.1021.acs.inorgchem.4c02394~\cite{wang_steerable_2024}      & 24 &  0 & 14 & 1.000 & 0.632 & 0.774 \\
10.1021.acs.inorgchem.8b01130~\cite{gosselin_design_2018}     &  7 &  0 & 25 & 1.000 & 0.219 & 0.359 \\
10.1021.acsami.7b09339~\cite{park_chromiumii_2017}            &  5 &  0 &  1 & 1.000 & 0.833 & 0.909 \\
10.1021.acsami.7b18836~\cite{lee_porous_2018}                 & 11 &  0 &  3 & 1.000 & 0.786 & 0.880 \\
10.1021.acsami.8b02015~\cite{barreda_ligand-based_2018}       &  7 &  0 &  4 & 1.000 & 0.636 & 0.778 \\
10.1021.cg4018322~\cite{paul_secondary_2014}                   &  5 &  0 & 14 & 1.000 & 0.263 & 0.417 \\
10.1021.ic050460z~\cite{ke_synthesis_2005}                     & 12 &  0 &  3 & 1.000 & 0.800 & 0.889 \\
10.1021.ic402428m~\cite{liu_situ_2013}                         & 28 &  0 &  5 & 1.000 & 0.848 & 0.918 \\
10.1021.ic501012e~\cite{tan_two_2014}                          & 18 &  0 & 12 & 1.000 & 0.600 & 0.750 \\
10.1021.ic802382p~\cite{prakash_metalorganic_2009}             &  7 &  0 &  7 & 1.000 & 0.500 & 0.667 \\
10.1021.ja042802q~\cite{sudik_design_2005}                     & 31 &  0 & 14 & 1.000 & 0.689 & 0.816 \\
10.1021.ja105986b~\cite{zheng_cubic_2010}                      &  6 &  0 &  5 & 1.000 & 0.545 & 0.706 \\
10.1021.jacs.7b00037~\cite{zhang_controlled_2017}              & 14 &  0 & 13 & 1.000 & 0.519 & 0.683 \\
10.1021.jacs.8b10866~\cite{zhang_self-assembly_2018}           & 30 &  0 & 29 & 1.000 & 0.508 & 0.674 \\
10.1039.C2CC34265K~\cite{du_giant_2012}                        & 17 &  0 & 15 & 1.000 & 0.531 & 0.694 \\
10.1039.C5CC05913E~\cite{augustyniak_vanadium_2015}            &  6 &  0 &  6 & 1.000 & 0.500 & 0.667 \\
10.1039.C5DT04764A~\cite{zhang_polyoxovanadate-based_2016}     & 12 &  0 & 26 & 1.000 & 0.316 & 0.480 \\
10.1039.C5RA26357C~\cite{chen_synthesis_2016}                  &  7 &  0 &  9 & 1.000 & 0.438 & 0.609 \\
10.1039.C6CC04583A~\cite{zhang_anderson-like_2016}             & 23 &  0 & 12 & 1.000 & 0.657 & 0.793 \\
10.1039.C6DT02764D~\cite{zhang_synthesis_2016}                 & 11 &  0 &  5 & 1.000 & 0.688 & 0.815 \\
10.1039.C7CC01208J~\cite{zhang_self-assembly_2017}             & 11 &  0 &  9 & 1.000 & 0.550 & 0.710 \\
10.1039.C8DT02580K~\cite{gong_functionalized_2018}             & 11 &  0 &  3 & 1.000 & 0.786 & 0.880 \\
10.1039.D3QI01501G~\cite{yang_efficient_2023}                  &  9 &  0 &  6 & 1.000 & 0.600 & 0.750 \\
\hline
Overall                                                          & 393 & 0 & 282 & 1.000 & 0.582 & 0.736 \\
\hline
\end{tabular}
}
\end{table*}

\subsubsection{Detailed instance visualisation for a UMC-1 synthesis}\label{subsubsec:umc1_instance_vis}

\begin{figure}[]
    \centering
    \includegraphics[angle=90, width=0.98\linewidth]{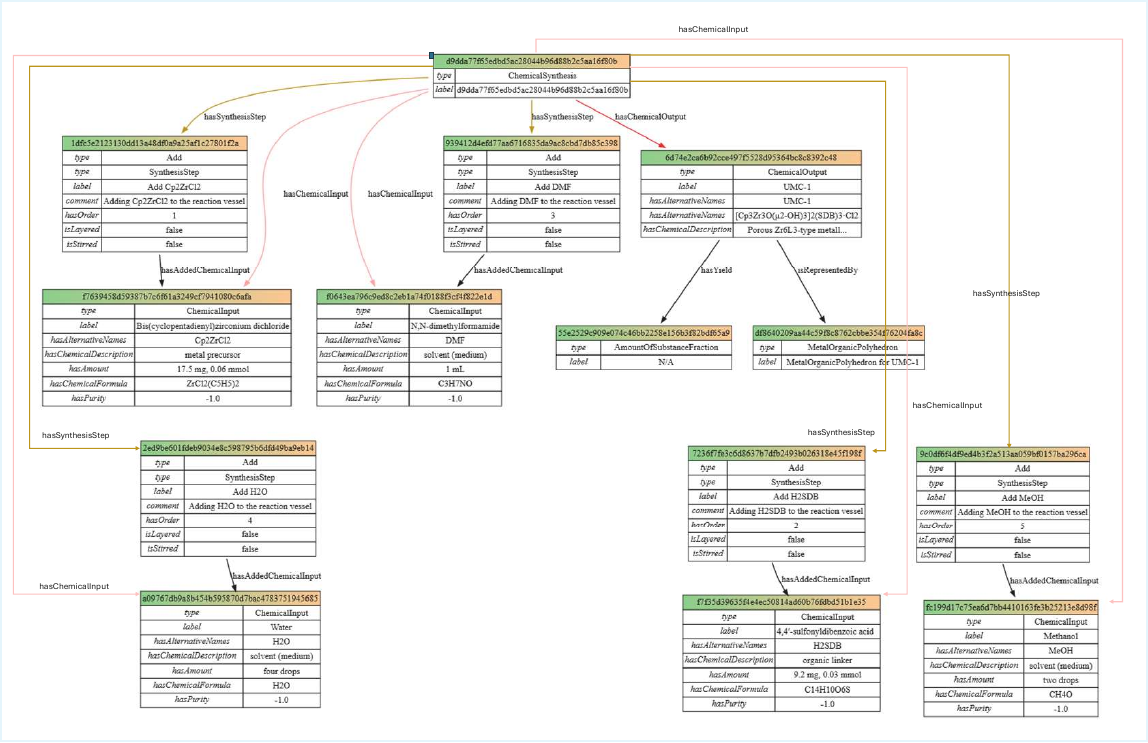}
    \caption{\textbf{Detailed ontology instance subgraph for a UMC-1 synthesis.}}
    \label{fig:umc1_instance_vis}
\end{figure}

The Supplementary Information includes a detailed visualisation of one instantiated ontology subgraph corresponding to the synthesis of the metal--organic polyhedron UMC-1 (Supplementary Fig.~\ref{fig:umc1_instance_vis}).
The figure shows the complete set of instantiated individuals and relations for this synthesis recipe under the OntoSyn and OntoMOPs ontologies, including typed synthesis steps, chemical inputs, and the resulting product entity, without the abstraction and condensation used in the main-text example (Fig.~\ref{fig:example_instance}).

The graph was initially rendered from the RDF serialisation using an RDF-to-Graphviz visualisation tool\footnote{\url{https://giacomociti.github.io/rdf2dot/}}
 and then manually edited to adjust the layout for presentation and page fitting.
\clearpage
\clearpage

\clearpage \citeindexfalse
\bibliography{bibliography}

\clearpage \makeciteindex

\end{document}